\documentclass[lettersize,journal]{IEEEtran}
\usepackage{amsmath,amsfonts}
\usepackage{algorithmic}
\usepackage{array}

\usepackage{textcomp}
\usepackage[authoryear]{natbib}
\usepackage{booktabs} 
\usepackage{stfloats}
\usepackage{multirow}
\usepackage{url}
\usepackage{subcaption} 
\usepackage{verbatim}
\usepackage{graphicx}
\def\BibTeX{{\rm B\kern-.05em{\sc i\kern-.025em b}\kern-.08em
    T\kern-.1667em\lower.7ex\hbox{E}\kern-.125emX}}
\usepackage{balance}

\begin{document}
\title{Aligning Human and Machine Attention for Enhanced Supervised Learning\thanks{This work has been submitted to the IEEE for possible publication. Copyright may be transferred without notice, after which this version may no longer be accessible.}}

\author{Avihay Chriqui, Inbal Yahav, Dov Te’eni, Ahmed Abbasi}


\maketitle

\date{October 2024}

\begin{abstract}
Attention, or prioritization of certain information items over others, is a critical element of any learning process, for both humans and machines. Given that humans continue to outperform machines in certain learning tasks, it seems plausible that machine performance could be enriched by aligning machine attention with human attention mechanisms---yet research on this topic is sparse and has achieved only limited success. This paper proposes a new approach to address this gap, called Human-Machine Attention Learning (HuMAL). This approach involves reliance on data annotated by humans to reflect their self-perceived attention during specific tasks. We evaluate several alternative strategies for integrating such human attention data into machine learning (ML) algorithms, using a sentiment analysis task (review data from Yelp) and a personality-type classification task (data from myPersonality). The best-performing HuMAL strategy significantly enhances the task performance of fine-tuned transformer models (BERT, as well as GPT-2 and XLNET), and the benefit is particularly pronounced under challenging conditions of imbalanced or sparse labeled data. This research contributes to a deeper understanding of strategies for integrating human attention into ML models and highlights the potential of leveraging human cognition to augment ML in real-world applications. 
\end{abstract}


\section{Introduction}

\IEEEPARstart{F}{or} many decision-making tasks, a collaboration between humans and machines, when realized effectively, is widely considered superior to either humans or machines working independently. Human-machine synergy is often achieved through various interactive paradigms, such as "human-in-the-loop" systems \citep{bhardwaj2023human,kim2023active,petricek2023ai} and reinforcement learning strategies \citep{te2023reciprocal}. This paper aims to further explore how human input can enrich machine learning processes, focusing on a crucial mechanism of learning: attention. 

Human attention involves the cognitive ability to concentrate on specific aspects of information while disregarding others. This selective focus is crucial for humans to manage the vast array of stimuli encountered daily, enabling effective task performance in areas like reading comprehension, text classification, and sentiment analysis. Yet human attention is not without its flaws; it can be biased and inconsistent and is often influenced by factors such as fatigue or cognitive overload \citep{yoo2022brain}.

Machine attention, in turn, derives from computational models that dynamically prioritize the most informative parts of input data, such as words or images \citep{luong2015effective,vania2017characters}. This principle of prioritizing parts of the available information, inspired by human attention strategies, has revolutionized deep learning models, improving both their performance and interpretability across various tasks in Natural Language Processing (NLP) and computer vision \citep{devlin2018bert,he2019using}. In particular, \citet{vaswani2017attention} introduced the transformer model, which relies entirely on attention mechanisms for sequence processing without recurrent or convolutional layers. Transformers use self-attention layers that allow each position in the input sequence to attend to all positions in the sequence, including itself. Self-attention allows the model to capture dependencies regardless of their distance in the input sequence, making it highly parallelizable and efficient for processing long sequences. 
Machine attention is more consistent than human attention and less prone to bias, and it scales effectively across large datasets. However, it is also limited by its dependency on training data and specific model architectures, which may hinder its ability to address novel or complex scenarios that human cognition handles more flexibly \citep{bhardwaj2023human}.

Given the suboptimal nature of current machine attention management, our objective is to enhance it by incorporating insights from human attention management. We anticipate that aligning machine attention with human attention will yield significant benefits, particularly in scenarios where the number of available training samples is limited across all or some of the classes, or in the absence of relevant context \citep{collins2018evolutionary, zagalsky2021design}.  To achieve this, we introduce a novel technique called Human-Machine Attention Learning (HuMAL), which leverages the attention patterns exhibited by humans, when performing specific tasks, to improve machine classification performance.

More specifically, the main research gaps we attempt to address are: (1) offering a deeper exposition of strategies for incorporating human attention into a transformer-based model; (2) proposing a method for leveraging human attention in machine learning that offers more effective performance; (3) providing an empirical analysis of the differential benefit provided by such a strategy as a function of class imbalance and labeled data availability.

\begin{figure}[!ht]
    \centering
    \resizebox{.7\columnwidth}{!}{%
    \includegraphics{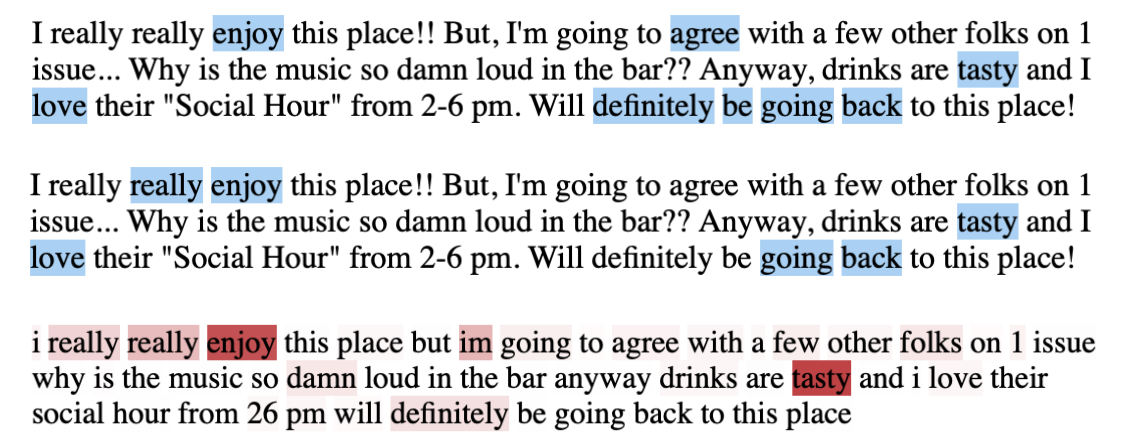}
    }
    \caption{Examples of human attention (blue in top two texts) compared with machine attention (red in bottom text) in a Yelp review. Adapted from ``Human Attention Maps for Text Classification: Do Humans and Neural Networks Focus on the Same Words?" by \citet{sen2020human}. Copyright 1963–2023 by ACL.}
    \label{fig:HuMAL_example_dataset}
\end{figure}

In what follows, we provide a brief overview of the distinctions between human and machine attention mechanisms and discuss prior efforts to integrate the two to improve machine learning performance. We then introduce our HuMAL approach, identifying different strategies to incorporate human attention into the attention mechanism of general-purpose large language models (LLMs), particularly BERT, XLNET, and GPT, for classification. We evaluate HuMAL, and the individual strategies specifically, in two tasks. The first is a sentiment analysis task, conducted on the Yelp review dataset provided by \citet{sen2020human}, which has previously been annotated for sentiment analysis purposes. Figure \ref{fig:HuMAL_example_dataset} (adapted from \citet[p. 4596]{sen2020human}) illustrates an example of annotated text from this dataset. The second is a personality type classification task, specifically, labeling texts (social media posts) as being written by introverts or extroverts. This task utilizes the well-known myPersonality social media dataset, gathered by \citet{stillwell2015mypersonality}, and which we annotated for human attention. In both tasks, following research that shows that human input is particularly beneficial when the availability of training data is limited \citep[e.g.,][]{zagalsky2021design}, we focus on the challenge of imbalanced text classification, with limited training instances. We show that incorporating human cognition into the machine attention mechanism provides an effective means of improving machine performance under such challenging conditions.

\section{Mechanisms of Human vs. Machine Attention in Learning Processes}

The concept of "human attention", as used herein, refers to how humans naturally focus on and give importance to different words in a text when reading or comprehending it. Human attention is based on cognitive processes and on the human brain's ability to process and interpret information, and it is influenced by past experiences and emotions. In the context of this paper, human attention is considered to be binary (in the sense that a human either attends to a particular aspect of the input or ignores it), with varying degrees of attention assigned to each word based on semantic and contextual cues. 

In contrast, attention in LLMs is continuous and probabilistic, computed based on learned representations and positional embeddings \citep{brauwers2021general}. Machine attention mechanisms, like those in BERT and GPT-3, focus on specific data parts by dynamically assigning weights to tokens. These mechanisms are effective for identifying patterns but are limited by training data and algorithms, resulting in a concrete and deterministic approach—unlike human attention \citep{brauwers2021general}, which can interpret abstract nuances and subtleties beyond explicit information, such as sarcasm or metaphors. Indeed, studies highlight that machine attention tends to be less flexible and more prone to focusing on spurious features, whereas human perception integrates abstract reasoning and context effectively, as seen in visual tasks and medical diagnoses \citep{lindsay2020attention, guo2021machine, makino2022differences}.

How similar or different are human attention and machine attention when reading texts? Previous research has examined this question from different perspectives and using different methods. For example, \citet{zou2023human} investigated whether humans and Deep Neural Networks (DNNs) allocate attention in comparable ways when reading a text passage to answer a specific question. They found that the DNN attention distribution quantitatively resembles human attention distribution measured by eye-tracking, and both are modulated by top-down reading goals and lower-level text features. 

The concept of \textit{attention correctness} entails using human attention to evaluate machine attention, and has been applied in the context of image caption text \citep{liu2017attention,brauwers2021general}. 
\citet{sen2020human} conducted a large-scale crowdsourcing study on a Yelp review dataset to collect human attention maps that encode the parts of a text that humans focus on when classifying texts. They compared human attention maps with machine attention maps created by DNNs and found that they had some overlap in word selections, but also some differences in distribution over lexical categories and context-dependency of sentiment polarity. 

\citet{cui2021understanding} analyzed how the performance of deep learning models is impacted by \textit{multi-head self-attention}: a core component of transformer models’ attention mechanism, in which the importance of each token is independently calculated by multiple submodules (``attention heads”), whose scores are subsequently combined \citep{vaswani2017attention, radford2019language, touvron2023llama}. Using eye-tracking data, the authors compared human and machine reading, highlighting the significant role of passage-to-question and passage-understanding attention. Their results also underscore the importance of certain word types and the influence of fine-tuning on attention distribution. 

\citet{zhao2023does} investigated if deep learning models can mimic human reading behavior, mainly focusing on significant words. The study discovers that while these models can indeed concentrate on significant words, they can also incorrectly focus on irrelevant ones due to inadequate learning. Thus, despite embodying some human reading traits, machine attention mechanisms exhibit certain limitations that necessitate more research. Finally, \citet{lei2022coca} explored the relationship between human and machine attention mechanisms in different neural network designs for computer vision tasks. The main finding of the paper was that the more closely artificial attention aligns with human attention, the better the model performance, suggesting that improving alignment may boost the performance of neural networks.

Given the research on the relative strengths of human attention compared with machine attention, we find little research on aligning machine mechanisms of attention with human attention. We found only a few reports on alignment designed to improve the performance or interpretability of deep learning models. For example, \citet{zou2018lexicon} proposed a way to use sentiment lexicons to guide the attention mechanism in deep learning models for sentiment classification tasks. They incorporated lexicon features into the attention vectors and regularized them with a lexicon-based loss function; their model outperformed previous methods on three large-scale sentiment classification datasets. However, the improvement was very small.

In a recent paper, \citet{mcguire2021sentiment} used both eye-tracking and brain imaging data (EEG) as ground truth to supervise the attention mechanism in BERT-based models, combining binary sentiment classification loss with attention losses. In this context, attention losses were computed by measuring the Kullback-Leibler (KL) divergence between (1) token-level EEG attention (average micro-voltage from 12 readers) and sentence-to-token machine attention, and (2) token-level eye-tracking attention (average reading time from 12 readers) and sentence-to-token machine attention. Notably, the readers in McGuire and Tomuro's paper were considered ``normal readers" rather than ``task-specific readers" \citep{hollenstein2018zuco}, implying that the attention they devoted to tokens was unaware of the specific sentiment task. The authors discovered that supervising machine attention with human attention could manipulate BERT's attention to be more human-like. Still, this manipulation did not result in significant differences in sentiment classification accuracy. Furthermore, they observed that models with cognitive attention supervision made different errors than baseline models and exhibited a higher ratio of false negatives.

Our goal in aligning machine attention with human attention to enhance machine learning and thereby boost task performance. Alignment may serve other purposes such as increasing explainability, which may affect performance. \citet{bao2018deriving} leveraged the concept of rationales to align machine attention with human attention. Model rationales refer to explanations or justifications for model predictions, often in the form of annotated subsets of data that are particularly relevant to the model’s decision. A neural network is designed with a rationale extraction module to identify and select important input segments sufficient for making predictions, which are then passed to a predictor module for the final classification decision\citep{lei2016rationalizing}. Bao and colleagues used human rationales to train the rationale extraction network module, before passing it to the classification. In the context of their paper, a human rationale is an annotated sentence from a given instance that is both necessary and sufficient to justify the labeler's decision. This method outperformed the examined baseline. 

However, the rationales approach has several limitations. Rationales typically involve post-hoc explanations that justify model predictions after the fact, which means they do not influence the model's learning process during training. Additionally, rationales often focus on a subset of input features, which can limit their ability to provide a comprehensive understanding of the entire input. 

We propose a technique that overcomes these limitations, integrating human attention directly into the model's attention mechanism during training, allowing for a more holistic alignment and potentially enhancing the model's performance.

\section{ HuMAL Strategies}

The HuMAL architecture is derived from a transformer model (e.g. encoder-decoder) for text classification, with modifications made to the attention layer that entail aligning the machine's attention with human attention.

The presentation of human attention data will follow this format: each token $i$ ($i = \{0, 1, ..., N\}$) in a sentence $j$, represented as $TOK_{j,i}$, will be associated with a human attention value $a^h_{j,i}$. This value is either binary, in the case of a single attention decision, or continuous in the case of multiple human attention decisions. 

For example, let us consider the last sentence in Figure \ref{fig:HuMAL_example_dataset}: ``Will definitely be going back to this place!". If we tokenize this sentence on a word level, we obtain the following tokens: [``will", ``definitely", ``be", ``going", ``back", ``to", ``this", ``place", ``!"]. The corresponding attention vector $A^h_j$ from the first (human) annotator would be [0, 1, 1, 1, 1, 0, 0, 0, 0]. 
Alternatively, if we tokenize the sentence into sub-words, it might look like this: [``will", ``def\#", ``\#initely", ``be", ``go\#", ``\#ing", ``back", ``to", ``this", ``place", ``!"]. In this case, when a human assigns attention to a word, we apply the attention decision to all subwords within that word. Thus, for the given example, the corresponding attention vector from the first annotator would be [0, 1, 1, 1, 1, 1, 1, 0, 0, 0, 0], indicating that attention is assigned to the subwords within the tokenized sentence accordingly.


In turn, the machine (e.g., BERT) attention value corresponding to a token $TOK_{j,i}$ is denoted $a^m_{j,i;l}$, where $l$ represents the transformer layer in the BERT architecture ($l=\{0, 1, ..., 11\}$). This attention value captures the focus that the model places on $TOK_i$ when processing the special token $CLS_j$ that represents the meaning of the entire sentence \citep{vaswani2017attention, devlin2018bert, clark2019does}. The value $a^m_{j,i;l}$ is computed via the average attention scores across all attention heads $h = \{0, 1, ..., 11\}$ within the same layer ($a^m_{j,i;l,h}$), given that heads within a layer have similar attention distributions \citep{clark2019does} (see Equation \ref{equ:machine_attention}). The machine attention vector for a given sentence $j$ is represented as the vector $A^m_{j;l}$.

\begin{equation}
    \label{equ:machine_attention}
    a^m_{j,i;l}  =  \frac{1}{12} \sum_{h}a^m_{j,i;l,h} 
\end{equation}

\begin{figure}[h]
    \centering
    \resizebox{\columnwidth}{!}{%
    \includegraphics{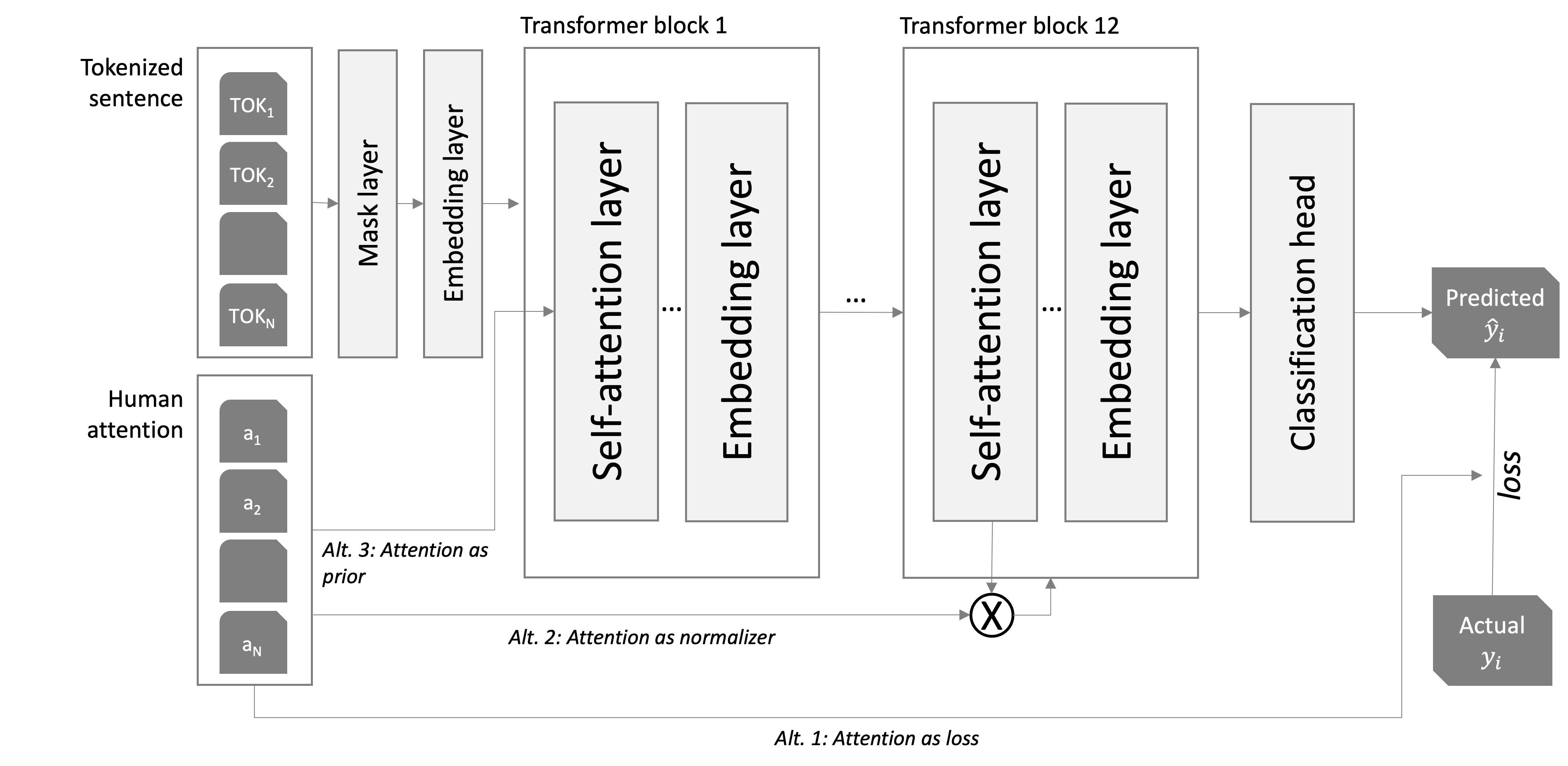}
    }
    \caption{HuMAL illustration: Alternative strategies for leveraging human attention into machine attention. Example using BERT as a base model}
    \label{fig:HuMAL}
\end{figure}

We explore three different strategies for aligning the machine's attention mechanism to human attention. All three strategies of HuMAL align machine attention to enhance classification performance, but they differ in the process of alignment. These strategies, illustrated in Figure \ref{fig:HuMAL} for a BERT model, are:

\begin{itemize}
    \item \textbf{Attention as loss (\textit{HuMAL-AL}):} 
    The model is trained to align its attention with human attention according to the loss function.  This method involves penalizing the model for deviating from human attention, in addition to the loss function with respect to the binary dependent value.    
    \item \textbf{Attention as normalizer (\textit{HuMAL-AN}):} With this approach, the human attention and the last learned attention layer are combined. The final sentence embedding representation takes into account both human and machine attention, allowing for the utilization of independent knowledge from each source.    
    \item \textbf{Attention as prior (\textit{HuMAL-AP}):} This method incorporates human attention as a prior in the first learned attention layer. By replacing the initial attention layer with human attention, the model is guided towards converging to a global optimum perceived by a human expert. 
\end{itemize}

The mathematical formulations of each approach are provided next.

\subsection{Attention as Loss (\textit{HuMAL-AL})}

\textit{HuMAL-AL} incorporates attention losses as a regularization technique for the LLM classification loss. Here, the classification loss, computed as the cross entropy between the sentence label $y_j$  and its predicted probability $p_j$, is regularized by the attention loss, which is calculated based on the cosine similarity between the human attention $A^h_j$ and the machine attention at the last layer $A^m_{j,11}$. The HuMAL-AL loss function is given in Equation \ref{equ:AL}.

\begin{equation}
    \begin{split}
    \label{equ:AL}
    AL\_loss =& \sum_j cross\_entropy(y_j,p_j) \\
    &+ \alpha \times cosine\_similarity(A^h_j, A^m_{j,11}) 
    \end{split}
\end{equation}

\subsection{Attention as Normalizer (\textit{HuMAL-AN})}

\textit{HuMAL-AN} applies normalization to the last layer of the model's attention, $A^m_{j,11}$, by using the human attention $A^h_j$. This normalization process influences the values of the last embedding vector of the $CLS_j$ token, represented as $E_{j,11}$. While this technique will not affect the training of the pre-trained model, adjusting the attention weights will immediately affect the level of importance assigned to the interaction between the tokens and the CLS tags. This, in turn, has implications for the classification task, which heavily relies on the final embedding representation. The HuMAL-AN modified embedding for sentence $j$ is given in Equation \ref{equ:AN}.

\begin{eqnarray}
    \label{equ:AN}
    E_{j,11}\_AN &=& (A^m_{j,11} + \alpha \times A^h_j) \times E_{j,10} \\
    & = & E_{j,11} + A^h_j \times E_{j,10} \nonumber
\end{eqnarray}

\subsection{Attention as Prior (\textit{HuMAL-AP})}

Similarly to \textit{HuMAL-AL}, \textit{HuMAL-AP} incorporates attention losses as a regularization technique for the LLM classification loss. However, there are key differences in how the losses are applied. Here, while cross entropy is employed in the last LLM layer, the attention loss is integrated into the first layer. The primary concept behind this alternative approach is to introduce human attention as a prior to the initial learned attention layer in LLM.

In the lower layers of LLM, especially BERT, attention heads tend to exhibit broader focus, allocating less than 10\% of their attention to individual words, resulting in a sentence representation that resembles a bag of vectors \citep{clark2019does}. By approximating the first attention layer with human attention, our aim is to enhance the model's likelihood of converging to a global optimum perceived by a human expert. This concept bears similarities to the utilization of manually defined centroids in unsupervised learning approaches.

The HuMAL-AP loss function is given in Equation \ref{equ:AP}.

\begin{equation}
    \begin{split}
        \label{equ:AP}
        AL\_loss = & \sum_j cross\_entropy(y_j,p_j) \\
        &+ \alpha \times cosine\_similarity(A^h_j, A^m_{j,0}) 
        \end{split}
\end{equation}

\section{Experiment}

To evaluate our model, we conducted experiments using two tasks: sentiment classification and personality type classification. The datasets utilized for these tasks are the Yelp dataset provided by \citet{sen2020human} and the myPersonality dataset from \citet{stillwell2015mypersonality}, respectively.

\subsection{Sentiment Classification Task}
For the sentiment classification, we used the dataset provided by \citet{sen2020human}. This dataset comprises Yelp restaurant reviews annotated for sentiment analysis, specifically identifying whether the sentiment expressed is positive or negative. 

To capture human attention, for each review in the dataset, \citet{sen2020human} asked crowd workers (three workers per review, on average) to highlight words they perceived as reflecting the review’s sentiment. These highlighted words serve as indicators of human attention directed towards the most significant words related to the overall task. The dataset is balanced and consists of approximately 4,500 instances. All sentiment labels were agreed upon by the annotators. 

It is worth noting that binary document-level sentiment polarity classification (as in the case of labeling positive versus negative reviews) is generally considered an ``easy" machine learning task. A BERT-based classification model, trained on a balanced subset of 2,000 instances from the dataset of \citet{sen2020human}, achieves a high AUC score of 0.98 (see more details in the Results section). 

We are interested in evaluating the performance of our HuMAL strategies on an imbalanced dataset, with few training instances. We thus undersampled the dataset, to adjust the ratio of the target class (either positive or negative sentiment) to 1, 2, 5, 10, and 20\% of the data, while varying the size of the training set between 250, 500, 1,000, and 2,000 instances. This procedure resulted in a total of 20 training subsets. Conversely, the test set remained balanced and consisted of a fixed number of 200 instances.

\subsection{Personality Type Classification Task}
For the personality type classification task, we used the myPersonality dataset by \citet{stillwell2015mypersonality}. MyPersonality is a large-scale research initiative to gather psychological data from Facebook users who consented to share their information. Participants completed personality surveys and provided access to their digital footprints (posts, likes, comments, etc.). The public dataset includes 250 users’ posts, in addition to self-reported personality type these users (i.e., introvert vs. extrovert); the full dataset became inaccessible in April 2018 due to the challenges associated with data management and ethical considerations. 

We sought to annotate and label users’ posts according to their personality types, namely, introvert vs. extrovert. To this end, we recruited workers from Prolific with an academic qualification in psychology and English as their native language. Our annotation procedure was similar to that employed by \citet{sen2020human}. Specifically, each instance underwent annotation by at least three annotators, with a preference for the first three annotators when the total number exceeded three. In each post, annotators indicated words they perceived as reflecting the writer’s personality type, and then labeled the personality type as either introvert or extrovert. Instances with less than 2\% of words annotated were excluded from the analysis.

Notably, we observed significant discrepancies between labels assigned by Prolific workers and those self-reported by users. In particular, users’ self-reported personality types aligned with Prolific workers’ assessments (majority vote) in only 124 out of 250 cases, as illustrated in Table \ref{tab:myPersonality_confusion_matrix}. 

\begin{table}[]
\centering
\caption{Confusion matrix comparison between the self-reported personality type and the majority vote of our annotators.}

\begin{tabular}{cccc}
\toprule
 & & \multicolumn{2}{c}{Majority Vote} \\
 \cmidrule{3-4} 
 & & Introvert & Extrovert \\
\multirow{2}{*}{Self Report} & 
\multicolumn{1}{c|}{Introvert} & 58 & 89 \\
 & \multicolumn{1}{c|}{Extrovert} & 26 & 66 \\ 
 \bottomrule
\end{tabular}
\label{tab:myPersonality_confusion_matrix}

\end{table}

Accordingly, for instances in which worker-assigned labels did not align with the self-reported label, we used two alternative approaches for including the instances’ corresponding annotations in our data: (i) including annotations concurring with the self-reported personality type; and (ii) including annotations agreeing with the majority vote. In each case, if there were more than three annotations that met the criteria, we used the first three. Table \ref{tab:HuMAL_myPersonality_count} shows the number of instances used for the training set in different experiment subsets.

\begin{table}[!]
\centering
\caption{The total training instances in different experiment subsets}

\begin{tabular}{lcc}
    \toprule
     & Imbalance (2-4\%) & Balance (40-60\%) \\
     \midrule
    Label by Majority & 105 & 160 \\
    Label by Self   Report & 54 & 107 \\
    \bottomrule
    \end{tabular}
\label{tab:HuMAL_myPersonality_count}

\end{table}

\subsection{Experiment Mechanism}

For each task and its corresponding dataset, we incorporated the attention data of all annotators to construct a continuous human attention vector. For instance, a token, \( a^h_{j,i}\) highlighted by three human annotators, \( t \in [0,3)\), receives three times more attention than a word highlighted by only one annotator \( a^h_{j,i} = \sum_{t=1}^{3}a^h_{j,i;t}\) . 

We normalized the sum of humans' attention vectors \( a^h\) to 1. 

\begin{equation}
    \label{equ:human_attention_mecha}
    \widetilde{a}^h_{j,i} = \frac{a^h_{j,i}}{\sum_{i=0}^{i}a^h_{j,i}}
\end{equation}

The human attention vectors were used only for training. 

While the human annotators focused on highlighting important words, the model's attention is distributed among sub-word tokens. Consequently, after training the model, we transferred the attention scores from tokens to words by aggregating the attention scores of tokens belonging to the same word.

We compared the performance of the three alternative HuMAL strategies across different imbalance ratios (for both tasks) and dataset sizes (for the sentiment analysis task only, as the dataset was much smaller for the personality type task, providing less flexibility). The experiments were conducted on a single GPU (A5000) with a batch size of 32. We set the learning rate to 5e-5 and trained the model for 10 epochs using the Adam optimizer. The regularization coefficient \(\alpha\) of all HuMAL approaches was fixed to 2. We use BERT-base-cased as a base model. To ensure reliable results, we repeated each experiment 20 times using the bootstrapping methodology.

\section{Results}

\subsection{Main Results: Comparing the Performance of the Three HuMAL Strategies}
\begin{figure}[!ht]
    \centering
    \resizebox{\columnwidth}{!}{%
    \includegraphics{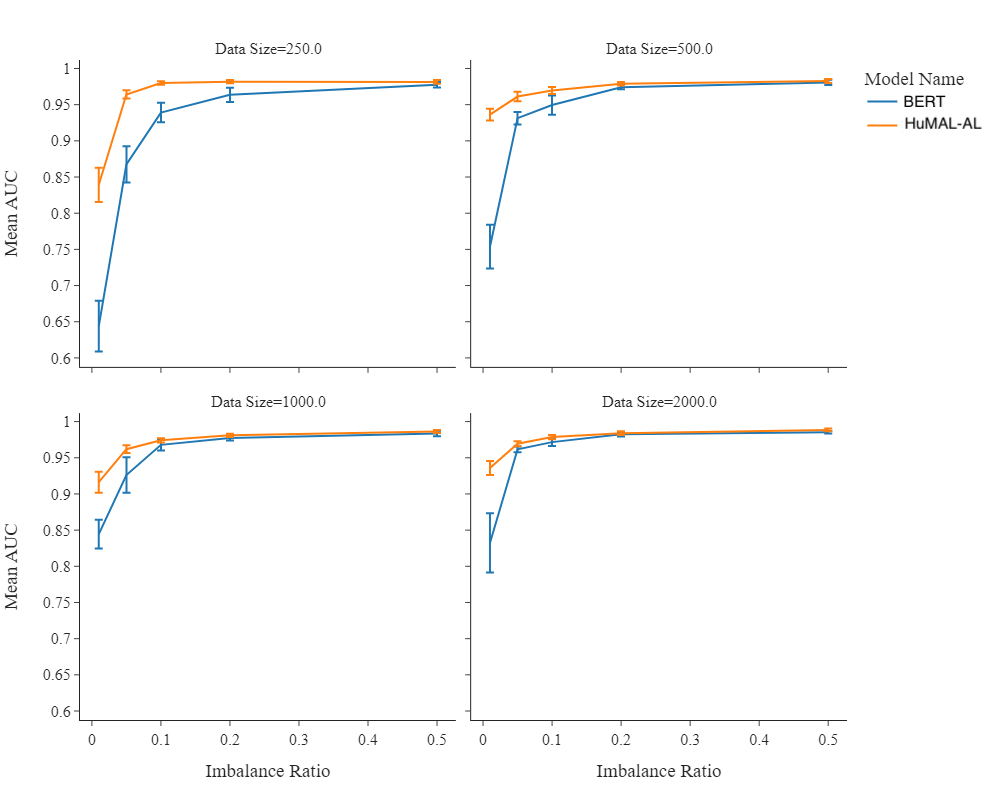}
    }
    \caption{Comparison between the performance of HuMAL-AL and BERT, as a function of data size and imbalance ratio, in the sentiment classification task.}
    \label{fig:HuMAL_AL_main}
\end{figure}

Table \ref{tab:HuMAL_perf} summarizes the performance of the HuMAL alternatives, compared to a general-purpose BERT model, fined-tuned to the specific task. Figure \ref{fig:HuMAL_AL_main} further illustrates the difference in performance between a BERT model and our first HuMAL strategy (\textit{HuMAL-AL}) on the sentiment classification task \footnote{We could not produce a similar figure for the personality task, due to the relatively small data size. }.

Overall, \textit{HuMAL-AL} demonstrated the best performance compared to both the baseline and the other HuMAL approaches. The performance benefit is particularly pronounced in challenging scenarios that contain a high class-imbalance ratio and a small number of training instances. The results were consistent across tasks (sentiment or personality type classification), regardless of the labeling of the minority class (e.g., ``positive" or ``negative" in sentiment analysis), and whether human attention was considered binary (first labeler only) or continuous. 
The other two approaches, \textit{HuMAL-AN} and \textit{HuMAL-AP}, showed a weak improvement over BERT, yet fell short compared to \textit{HuMAL-AL}. Together, these results provide evidence of the benefit of leveraging human attention---particularly through the \textit{HuMAL-AL} approach---to improve machine learning performance. In what follows, we further characterize various aspects of \textit{HuMAL-AL}’s performance.

\begin{table*}[!ht]
\caption{Performance of alternative HuMAL strategies}
\label{tab:HuMAL_perf}
\begin{subtable}{\textwidth}
\centering

\subcaption{Sentiment analysis classification}
\begin{tabular}{lcccccccccc}
\toprule
 &
  \multicolumn{5}{c}{Dataset   Size = 250} &
  \multicolumn{5}{c}{Dataset Size = 500} \\
 &
  1\% &
  5\% &
  10\% &
  20\% &
  50\% &
  1\% &
  5\% &
  10\% &
  20\% &
  50\% \\ \midrule
BERT &
  \begin{tabular}[c]{@{}c@{}}0.64\\      (0.11)\end{tabular} &
  \begin{tabular}[c]{@{}c@{}}0.87\\      (0.08)\end{tabular} &
  \begin{tabular}[c]{@{}c@{}}0.94\\      (0.04)\end{tabular} &
  \begin{tabular}[c]{@{}c@{}}0.96\\      (0.03)\end{tabular} &
  \begin{tabular}[c]{@{}c@{}}0.98\\      (0.01)\end{tabular} &
  \begin{tabular}[c]{@{}c@{}}0.75\\      (0.1)\end{tabular} &
  \begin{tabular}[c]{@{}c@{}}0.93\\      (0.03)\end{tabular} &
  \begin{tabular}[c]{@{}c@{}}0.95\\      (0.04)\end{tabular} &
  \begin{tabular}[c]{@{}c@{}}0.97\\      (0.01)\end{tabular} &
  \begin{tabular}[c]{@{}c@{}}0.98\\      (0.01)\end{tabular} \\
HuMAL-AL &
  \begin{tabular}[c]{@{}c@{}}0.84*\\      (0.08)\end{tabular} &
  \begin{tabular}[c]{@{}c@{}}0.96*\\      (0.02)\end{tabular} &
  \begin{tabular}[c]{@{}c@{}}0.98*\\      (0.01)\end{tabular} &
  \begin{tabular}[c]{@{}c@{}}0.98*\\      (0.01)\end{tabular} &
  \begin{tabular}[c]{@{}c@{}}0.98\\      (0.01)\end{tabular} &
  \begin{tabular}[c]{@{}c@{}}0.94*\\      (0.03)\end{tabular} &
  \begin{tabular}[c]{@{}c@{}}0.96*\\      (0.02)\end{tabular} &
  \begin{tabular}[c]{@{}c@{}}0.97\\      (0.02)\end{tabular} &
  \begin{tabular}[c]{@{}c@{}}0.98\\      (0.01)\end{tabular} &
  \begin{tabular}[c]{@{}c@{}}0.98\\      (0.01)\end{tabular} \\
HuMAL-AN &
  \begin{tabular}[c]{@{}c@{}}0.69\\      (0.13)\end{tabular} &
  \begin{tabular}[c]{@{}c@{}}0.9\\      (0.04)\end{tabular} &
  \begin{tabular}[c]{@{}c@{}}0.95\\      (0.03)\end{tabular} &
  \begin{tabular}[c]{@{}c@{}}0.97\\      (0.01)\end{tabular} &
  \begin{tabular}[c]{@{}c@{}}0.98\\      (0.01)\end{tabular} &
  \begin{tabular}[c]{@{}c@{}}0.79\\      (0.06)\end{tabular} &
  \begin{tabular}[c]{@{}c@{}}0.95*\\      (0.02)\end{tabular} &
  \begin{tabular}[c]{@{}c@{}}0.97\\      (0.01)\end{tabular} &
  \begin{tabular}[c]{@{}c@{}}0.97\\      (0.01)\end{tabular} &
  \begin{tabular}[c]{@{}c@{}}0.98\\      (0.01)\end{tabular} \\
HuMAL-AP &
  \begin{tabular}[c]{@{}c@{}}0.7\\      (0.1)\end{tabular} &
  \begin{tabular}[c]{@{}c@{}}0.91*\\      (0.04)\end{tabular} &
  \begin{tabular}[c]{@{}c@{}}0.95\\      (0.04)\end{tabular} &
  \begin{tabular}[c]{@{}c@{}}0.97\\      (0.02)\end{tabular} &
  \begin{tabular}[c]{@{}c@{}}0.98\\      (0.01)\end{tabular} &
  \begin{tabular}[c]{@{}c@{}}0.79\\      (0.06)\end{tabular} &
  \begin{tabular}[c]{@{}c@{}}0.94\\      (0.04)\end{tabular} &
  \begin{tabular}[c]{@{}c@{}}0.95\\      (0.04)\end{tabular} &
  \begin{tabular}[c]{@{}c@{}}0.98\\      (0.01)\end{tabular} &
  \begin{tabular}[c]{@{}c@{}}0.98\\      (0.01)\end{tabular} \\

 &
   &
   &
   &
   &
   &
   &
   &
   &
   &
   \\
 &
  \multicolumn{5}{c}{Dataset Size = 1000} &
  \multicolumn{5}{c}{Dataset Size = 2000} \\ 
 &
  1\% &
  5\% &
  10\% &
  20\% &
  50\% &
  1\% &
  5\% &
  10\% &
  20\% &
  50\% \\ \midrule
BERT &
  \begin{tabular}[c]{@{}c@{}}0.84\\      (0.06)\end{tabular} &
  \begin{tabular}[c]{@{}c@{}}0.93\\      (0.08)\end{tabular} &
  \begin{tabular}[c]{@{}c@{}}0.97\\      (0.02)\end{tabular} &
  \begin{tabular}[c]{@{}c@{}}0.98\\      (0.01)\end{tabular} &
  \begin{tabular}[c]{@{}c@{}}0.98\\      (0.01)\end{tabular} &
  \begin{tabular}[c]{@{}c@{}}0.83\\      (0.13)\end{tabular} &
  \begin{tabular}[c]{@{}c@{}}0.96\\      (0.01)\end{tabular} &
  \begin{tabular}[c]{@{}c@{}}0.97\\      (0.02)\end{tabular} &
  \begin{tabular}[c]{@{}c@{}}0.98\\      (0.01)\end{tabular} &
  \begin{tabular}[c]{@{}c@{}}0.99\\      (0.01)\end{tabular} \\
HuMAL-AL &
  \begin{tabular}[c]{@{}c@{}}0.92*\\      (0.05)\end{tabular} &
  \begin{tabular}[c]{@{}c@{}}0.96\\      (0.02)\end{tabular} &
  \begin{tabular}[c]{@{}c@{}}0.97\\      (0.01)\end{tabular} &
  \begin{tabular}[c]{@{}c@{}}0.98\\      (0.01)\end{tabular} &
  \begin{tabular}[c]{@{}c@{}}0.99\\      (0.01)\end{tabular} &
  \begin{tabular}[c]{@{}c@{}}0.94*\\      (0.03)\end{tabular} &
  \begin{tabular}[c]{@{}c@{}}0.97\\      (0.01)\end{tabular} &
  \begin{tabular}[c]{@{}c@{}}0.98\\      (0.01)\end{tabular} &
  \begin{tabular}[c]{@{}c@{}}0.98\\      (0.01)\end{tabular} &
  \begin{tabular}[c]{@{}c@{}}0.99\\      (0.01)\end{tabular} \\
HuMAL-AN &
  \begin{tabular}[c]{@{}c@{}}0.84\\      (0.09)\end{tabular} &
  \begin{tabular}[c]{@{}c@{}}0.96\\      (0.02)\end{tabular} &
  \begin{tabular}[c]{@{}c@{}}0.97\\      (0.01)\end{tabular} &
  \begin{tabular}[c]{@{}c@{}}0.98\\      (0.01)\end{tabular} &
  \begin{tabular}[c]{@{}c@{}}0.98\\      (0.01)\end{tabular} &
  \begin{tabular}[c]{@{}c@{}}0.87\\      (0.09)\end{tabular} &
  \begin{tabular}[c]{@{}c@{}}0.97\\      (0.01)\end{tabular} &
  \begin{tabular}[c]{@{}c@{}}0.97\\      (0.01)\end{tabular} &
  \begin{tabular}[c]{@{}c@{}}0.98\\      (0.01)\end{tabular} &
  \begin{tabular}[c]{@{}c@{}}0.98\\      (0.01)\end{tabular} \\
HuMAL-AP &
  \begin{tabular}[c]{@{}c@{}}0.8\\      (0.13)\end{tabular} &
  \begin{tabular}[c]{@{}c@{}}0.96*\\      (0.02)\end{tabular} &
  \begin{tabular}[c]{@{}c@{}}0.97\\      (0.01)\end{tabular} &
  \begin{tabular}[c]{@{}c@{}}0.98\\      (0.01)\end{tabular} &
  \begin{tabular}[c]{@{}c@{}}0.99\\      (0.01)\end{tabular} &
  \begin{tabular}[c]{@{}c@{}}0.9*\\      (0.04)\end{tabular} &
  \begin{tabular}[c]{@{}c@{}}0.97*\\      (0.01)\end{tabular} &
  \begin{tabular}[c]{@{}c@{}}0.97\\      (0.02)\end{tabular} &
  \begin{tabular}[c]{@{}c@{}}0.98\\      (0.01)\end{tabular} &
  \begin{tabular}[c]{@{}c@{}}0.99\\      (0.01)\end{tabular} \\

  \bottomrule
\end{tabular}

\end{subtable}
 \vspace{0.5cm} 
 
 \centering

\begin{subtable}{\textwidth}
\centering
\subcaption{Personality type classification}
\centering
\begin{tabular}{lcccc}
\toprule
& \multicolumn{2}{c}{Label by Majority} & \multicolumn{2}{c}{Label by Self Report} \\ 
 & Imbalance & Balance & Imbalance & Balance \\
\midrule
BERT & \begin{tabular}[c]{@{}c@{}}0.57\\ (0.11)\end{tabular} & \begin{tabular}[c]{@{}c@{}}0.68\\ (0.09)\end{tabular} & \begin{tabular}[c]{@{}c@{}}0.60 \\ (0.09)\end{tabular} & \begin{tabular}[c]{@{}c@{}}0.64\\ (0.08)\end{tabular} \\
HuMAL-AL & \begin{tabular}[c]{@{}c@{}}0.59\\ (0.09)\end{tabular} & \begin{tabular}[c]{@{}c@{}}0.74*\\ (0.07)\end{tabular} & \begin{tabular}[c]{@{}c@{}}0.62\\ (0.11)\end{tabular} & \begin{tabular}[c]{@{}c@{}}0.75*\\ (0.08)\end{tabular} \\

HuMAL-AN & \begin{tabular}[c]{@{}c@{}}0.54\\ (0.13)\end{tabular} & \begin{tabular}[c]{@{}c@{}}0.56\\ (0.07)\end{tabular} & \begin{tabular}[c]{@{}c@{}}0.52\\ (0.08)\end{tabular} & \begin{tabular}[c]{@{}c@{}}0.55\\ (0.09)\end{tabular} \\

HuMAL-AP & \begin{tabular}[c]{@{}c@{}}0.58\\ (0.09)\end{tabular} & \begin{tabular}[c]{@{}c@{}}0.65\\ (0.10)\end{tabular} & \begin{tabular}[c]{@{}c@{}}0.60\\ (0.09)\end{tabular} & \begin{tabular}[c]{@{}c@{}}0.65\\ (0.06)\end{tabular} \\

\bottomrule
\end{tabular}

\end{subtable}

\tiny
The results are the mean of 20 random bootstraps. * denotes a significant improvement over BERT model
\end{table*}

\subsection{Data Requirements of HuMAL-AL vs. BERT}

\begin{figure}[!ht]
    \centering
    \resizebox{.8\columnwidth}{!}{%
    \includegraphics[width=.5\textwidth]{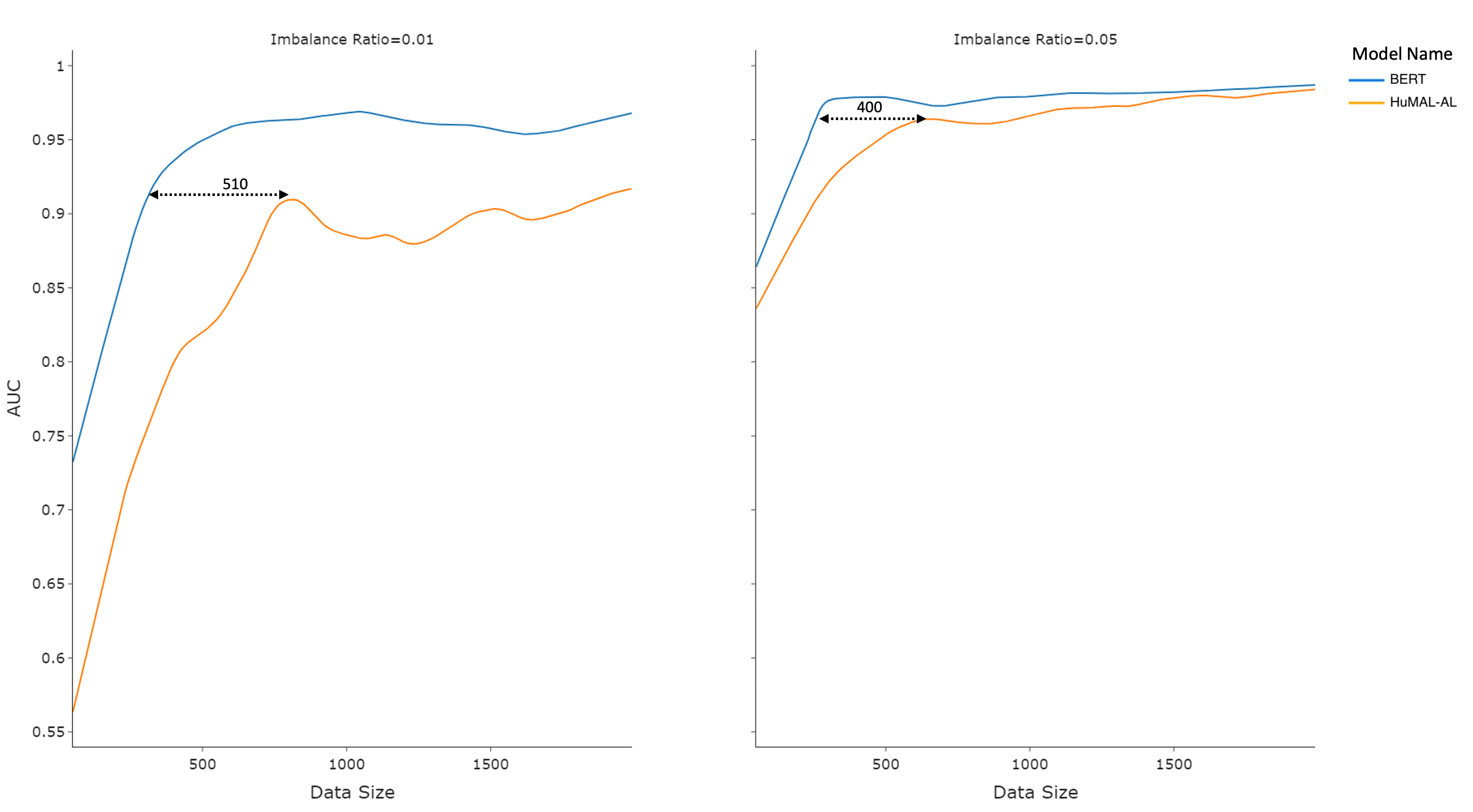}
    }
    \caption{Performance of HuMAL-AL compared to BERT across different dataset sizes in the sentiment classification task.}
    \label{fig:label_cost_alt3}
\end{figure}

We conducted a comprehensive analysis to compare the amount of labeled data required by a BERT model versus \textit{HuMAL-AL} while controlling for performance (AUC); this analysis provides an assessment of the extent to which HuMAL enables users to save on labeling costs. We focused on the sentiment labeling task only, as the personality type dataset did not contain enough instances for the analysis. Specifically, we varied the size of the training set from 50 to 2000 observations, maintaining a constant imbalance ratio. The two extreme imbalance ratios considered were 1\% and 5\%. Figure \ref{fig:label_cost_alt3} presents a visualization of the results of this analysis, where the AUC curves are displayed using a lowess function with frac=0.2 applied to the AUC values.

Overall, the results demonstrate that HuMAL requires significantly fewer training observations than BERT does to achieve comparable performance. When the imbalance ratio is 5\%, the average reduction in required labeled data amounts to 484 instances. When the imbalance ratio is higher (1\%), BERT requires an average of 780 labeled instances to match HuMAL's performance, provided that the AUC is lower than ~0.9. For higher AUC values, BERT fails to attain the performance of HuMAL, regardless of the number of labeled instances.

\subsection{Impact on the Classification Loss}

\begin{figure}[!ht]
    \centering
    \begin{minipage}{\columnwidth}
        \centering
        \includegraphics[width=\linewidth]{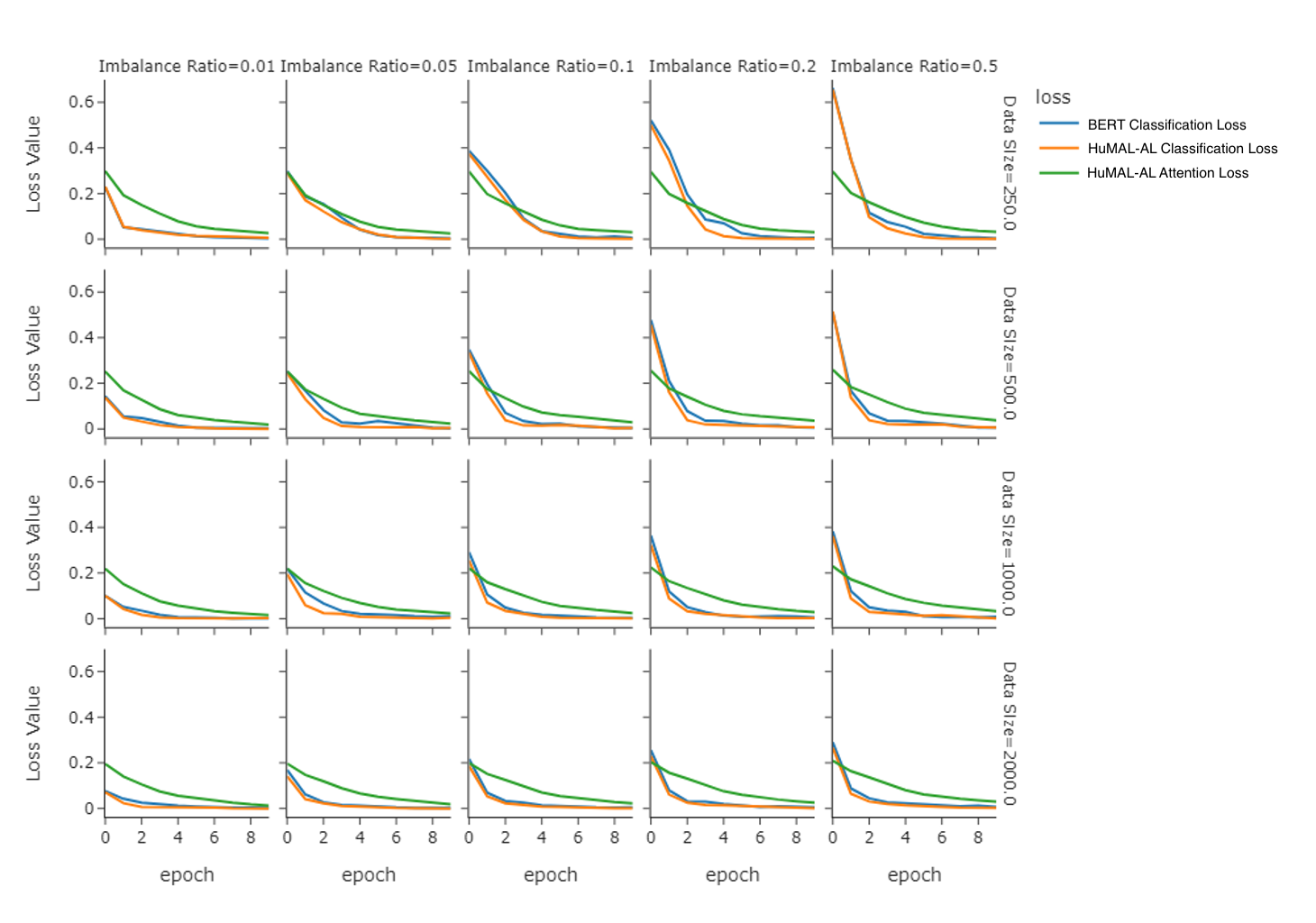}
        \subcaption{Loss Convergence Analysis for Sentiment Classification}
    \end{minipage}
    
    \begin{minipage}{\columnwidth}
        \centering
        \includegraphics[width=\linewidth]{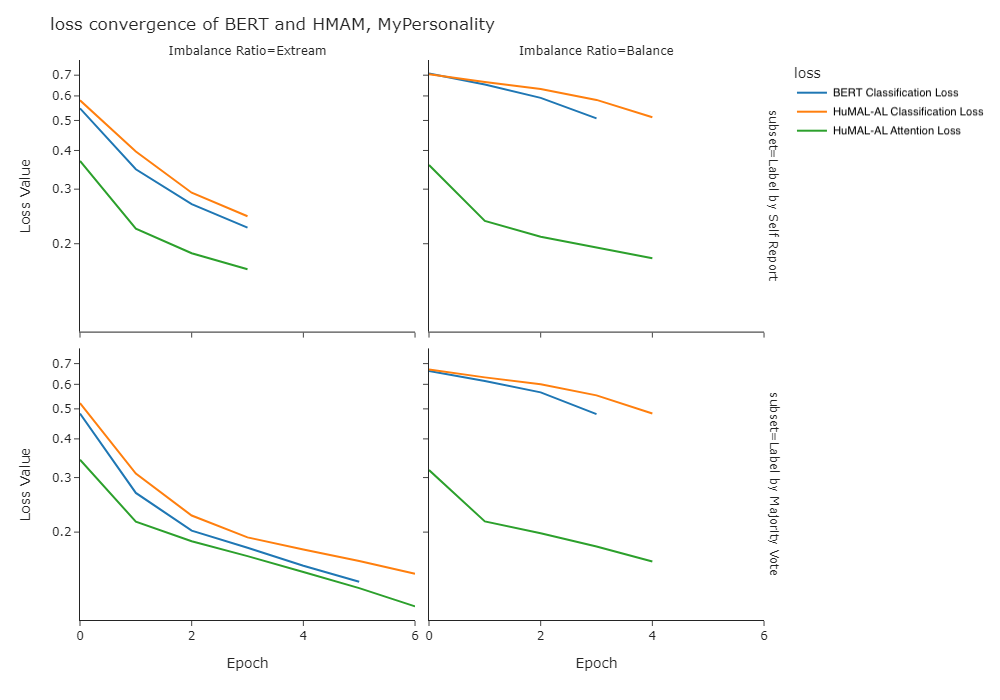}
        \subcaption{Loss Convergence Analysis for Personality Type Classification}
    \end{minipage}
    \caption{HuMAL-AL: Loss Convergence Analysis}
    \label{fig:alt3_loss_analysis}
\end{figure}

To understand how incorporating human attention affects the model's training process, we analyzed (for both tasks) the convergence of the \textit{HuMAL-AL} loss function. We observed that HuMAL required more epochs to converge compared to BERT. However, it eventually reached the same overall loss as BERT without compromising attention or classification accuracy. By decomposing the HuMAL loss function, as shown in Figure \ref{fig:alt3_loss_analysis}, we discovered that the classification loss (cross-entropy) of HuMAL behaves similarly to BERT's throughout all the epochs. In parallel, the attention loss of HuMAL converges monotonously without any tradeoff. This observation highlights the effectiveness of HuMAL in enhancing attention modeling while preserving classification performance.

\subsection{Impact of Text Length on HuMAL Performance}

We investigated the performance of the HuMAL across texts of different lengths, taking into account the insights provided by \citet{sen2020human} regarding the decrease in agreement among human annotators as text length increases, while classification decisions (i.e., negative/positive) remained unchanged. This result suggests that human attention data become less informative as the text length increases. We thus anticipated that, for texts of greater length, \textit{HuMAL-AL} would provide less pronounced improvement compared to BERT. 

For the sentiment classification task, we divided the texts in the Yelp dataset into three categories based on length, in accordance with \citet{sen2020human}: (1) up to 50 words (denoted Yelp-50), (2) 51-100 words (Yelp-100), and (3) 101-200 words (Yelp-200). As expected, HuMAL's improvement over BERT decreased with text length (see Table \ref{tab:alt3_text_length}), reflecting the decreased in agreement among annotators. Nevertheless, HuMAL consistently demonstrated robust performance across all text lengths, outperforming BERT. 

For the personality type classification task, we split the myPersonality dataset into two groups based on text length: short texts (up to 100 words) and longer texts (101 words and above). Similarly to what we observed in the Yelp analysis, we found that the improvement offered by HuMAL decreased with increasing text length. Specifically, the shorter the text, the greater the enhancement achieved by HuMAL. Nevertheless, HuMAL consistently outperformed BERT across both short and long texts. 

These observations reinforce the effectiveness of HuMAL in capturing and leveraging human attention cues, particularly in contexts where text length may influence the agreement among annotators and the complexity of attention modeling.

\begin{table*}[!ht]
\caption{HuMAL performance for text instances of different lengths}
\label{tab:alt3_text_length}

\centering
\begin{subtable}{\textwidth}
\centering
\subcaption{Yelp}
\begin{tabular}{cccccccccccc}
\toprule
 &
   &
  \multicolumn{5}{c}{Data Size = 250} &
  \multicolumn{5}{c}{Data Size = 500} \\
 &
   &
  1\% &
  5\% &
  10\% &
  20\% &
  50\% &
  1\% &
  5\% &
  10\% &
  20\% &
  50\% \\ \midrule
Yelp-50 &
  BERT &
  \begin{tabular}[c]{@{}c@{}}0.68\\      (0.13)\end{tabular} &
  \begin{tabular}[c]{@{}c@{}}0.87\\      (0.06)\end{tabular} &
  \begin{tabular}[c]{@{}c@{}}0.92\\      (0.07)\end{tabular} &
  \begin{tabular}[c]{@{}c@{}}0.96\\      (0.01)\end{tabular} &
  \begin{tabular}[c]{@{}c@{}}0.97\\      (0.01)\end{tabular} &
  \begin{tabular}[c]{@{}c@{}}0.73\\      (0.11)\end{tabular} &
  \begin{tabular}[c]{@{}c@{}}0.9\\      (0.07)\end{tabular} &
  \begin{tabular}[c]{@{}c@{}}0.95\\      (0.03)\end{tabular} &
  \begin{tabular}[c]{@{}c@{}}0.96\\      (0.03)\end{tabular} &
  \begin{tabular}[c]{@{}c@{}}0.98\\      (0.01)\end{tabular} \\
 &
  HuMAL-AL &
  \begin{tabular}[c]{@{}c@{}}0.82*\\      (0.07)\end{tabular} &
  \begin{tabular}[c]{@{}c@{}}0.95*\\      (0.02)\end{tabular} &
  \begin{tabular}[c]{@{}c@{}}0.97*\\      (0.01)\end{tabular} &
  \begin{tabular}[c]{@{}c@{}}0.98*\\      (0.01)\end{tabular} &
  \begin{tabular}[c]{@{}c@{}}0.98*\\      (0.01)\end{tabular} &
  \begin{tabular}[c]{@{}c@{}}0.92*\\      (0.03)\end{tabular} &
  \begin{tabular}[c]{@{}c@{}}0.95*\\      (0.03)\end{tabular} &
  \begin{tabular}[c]{@{}c@{}}0.95\\      (0.04)\end{tabular} &
  \begin{tabular}[c]{@{}c@{}}0.97\\      (0.02)\end{tabular} &
  \begin{tabular}[c]{@{}c@{}}0.98\\      (0.01)\end{tabular} \\
Yelp-100 &
  BERT &
  \begin{tabular}[c]{@{}c@{}}0.64\\      (0.1)\end{tabular} &
  \begin{tabular}[c]{@{}c@{}}0.9\\      (0.07)\end{tabular} &
  \begin{tabular}[c]{@{}c@{}}0.95\\      (0.03)\end{tabular} &
  \begin{tabular}[c]{@{}c@{}}0.98\\      (0.01)\end{tabular} &
  \begin{tabular}[c]{@{}c@{}}0.98\\      (0.01)\end{tabular} &
  \multicolumn{3}{c}{\multirow{2}{*}{\textit{Not enough   data}}} &
  \begin{tabular}[c]{@{}c@{}}0.98\\      (0.01)\end{tabular} &
  \begin{tabular}[c]{@{}c@{}}0.99\\      (0.01)\end{tabular} \\
 &
  HuMAL-AL &
  \begin{tabular}[c]{@{}c@{}}0.83*\\      (0.1)\end{tabular} &
  \begin{tabular}[c]{@{}c@{}}0.96*\\      (0.01)\end{tabular} &
  \begin{tabular}[c]{@{}c@{}}0.98*\\      (0.01)\end{tabular} &
  \begin{tabular}[c]{@{}c@{}}0.98\\      (0.01)\end{tabular} &
  \begin{tabular}[c]{@{}c@{}}0.98\\      (0.01)\end{tabular} &
  \multicolumn{3}{c}{} &
  \begin{tabular}[c]{@{}c@{}}0.98*\\      (0.01)\end{tabular} &
  \begin{tabular}[c]{@{}c@{}}0.99\\      (0.01)\end{tabular} \\
 &
   &
   &
   &
   &
   &
   &
   &
   &
   &
   &
   \\
 &
   &
  \multicolumn{5}{c}{Data Size = 1000} &
  \multicolumn{5}{c}{Data Size = 2000} \\
 &
   &
  1\% &
  5\% &
  10\% &
  20\% &
  50\% &
  1\% &
  5\% &
  10\% &
  20\% &
  50\% \\ \midrule
Yelp-50 &
  BERT &
  \begin{tabular}[c]{@{}c@{}}0.8\\      (0.1)\end{tabular} &
  \begin{tabular}[c]{@{}c@{}}0.95\\      (0.02)\end{tabular} &
  \begin{tabular}[c]{@{}c@{}}0.96\\      (0.02)\end{tabular} &
  \begin{tabular}[c]{@{}c@{}}0.98\\      (0.01)\end{tabular} &
  \begin{tabular}[c]{@{}c@{}}0.98\\      (0.01)\end{tabular} &
  \multicolumn{4}{c}{\multirow{2}{*}{\textit{Not enough data}}} &
  \begin{tabular}[c]{@{}c@{}}0.98\\      (0.01)\end{tabular} \\
 &
  HuMAL-AL &
  \begin{tabular}[c]{@{}c@{}}0.91*\\      (0.1)\end{tabular} &
  \begin{tabular}[c]{@{}c@{}}0.95\\      (0.05)\end{tabular} &
  \begin{tabular}[c]{@{}c@{}}0.97\\      (0.02)\end{tabular} &
  \begin{tabular}[c]{@{}c@{}}0.98\\      (0.01)\end{tabular} &
  \begin{tabular}[c]{@{}c@{}}0.98\\      (0.01)\end{tabular} &
  \multicolumn{4}{c}{} &
  \begin{tabular}[c]{@{}c@{}}0.99*\\      (0.01)\end{tabular} \\
Yelp-100 &
  BERT &
  \multicolumn{10}{c}{\multirow{2}{*}{\textit{Not enough   data}}} \\
 &
  HuMAL-AL &
  \multicolumn{10}{c}{} \\
 &
   &
   &
   &
   &
   &
   &
   &
   &
   &
   &
   \\
 &
   &
  \multicolumn{5}{c}{Data Size = 50} &
  \multicolumn{5}{c}{Data Size = 100} \\
 &
   &
  1\% &
  5\% &
  10\% &
  20\% &
  50\% &
  1\% &
  5\% &
  10\% &
  20\% &
  50\% \\ \midrule
Yelp-200 &
  BERT &
  \begin{tabular}[c]{@{}c@{}}0.53\\      (0.18)\end{tabular} &
  \begin{tabular}[c]{@{}c@{}}0.65\\      (0.2)\end{tabular} &
  \begin{tabular}[c]{@{}c@{}}0.84\\      (0.09)\end{tabular} &
  \begin{tabular}[c]{@{}c@{}}0.9\\      (0.05)\end{tabular} &
  \begin{tabular}[c]{@{}c@{}}0.93\\      (0.11)\end{tabular} &
  \multicolumn{4}{c}{\multirow{2}{*}{\textit{Not enough}}} &
  \begin{tabular}[c]{@{}c@{}}0.95\\      (0.05)\end{tabular} \\
Yelp-200 &
  HuMAL-AL &
  \begin{tabular}[c]{@{}c@{}}0.63*\\      (0.12)\end{tabular} &
  \begin{tabular}[c]{@{}c@{}}0.76*\\      (0.11)\end{tabular} &
  \begin{tabular}[c]{@{}c@{}}0.85\\      (0.1)\end{tabular} &
  \begin{tabular}[c]{@{}c@{}}0.92\\      (0.05)\end{tabular} &
  \begin{tabular}[c]{@{}c@{}}0.95\\      (0.04)\end{tabular} &
  \multicolumn{4}{c}{} &
  \begin{tabular}[c]{@{}c@{}}0.97\\      (0.02)\end{tabular}
\\
\bottomrule

\end{tabular}
\end{subtable}

 \vspace{0.5cm} 
\centering

\begin{subtable}{\textwidth}
\subcaption{MyPersonality}
\centering
\begin{tabular}{cccccc}
\toprule
\textbf{} & \textbf{} & \multicolumn{2}{c}{Label by Majority} & \multicolumn{2}{c}{Label by Self Report} \\
    \textbf{} & \textbf{} & Imbalance & Balance & Imbalance & Balance \\
\midrule
\multirow{4}{*}{\begin{tabular}[c]{@{}c@{}}Text\\ Length\\ \textless{}100\end{tabular}} & BERT & \begin{tabular}[c]{@{}c@{}}0.54\\ (0.14)\end{tabular} & \begin{tabular}[c]{@{}c@{}}0.7\\ (0.09)\end{tabular} & \begin{tabular}[c]{@{}c@{}}0.59\\ (0.14)\end{tabular} & \begin{tabular}[c]{@{}c@{}}0.61\\ (0.14)\end{tabular} \\
 & HuMAL-AL & \begin{tabular}[c]{@{}c@{}}0.64*\\ (0.13)\end{tabular} & \begin{tabular}[c]{@{}c@{}}0.77*\\ (0.09)\end{tabular} & \begin{tabular}[c]{@{}c@{}}0.67*\\ (0.14)\end{tabular} & \begin{tabular}[c]{@{}c@{}}0.68\\ (0.13)\end{tabular} \\

\multirow{4}{*}{\begin{tabular}[c]{@{}c@{}}Text\\ Length\\ \textgreater{}100\end{tabular}} & BERT & \begin{tabular}[c]{@{}c@{}}0.54\\ (0.12)\end{tabular} & \begin{tabular}[c]{@{}c@{}}0.69\\ (0.09)\end{tabular} & \begin{tabular}[c]{@{}c@{}}0.55\\ (0.16)\end{tabular} & \begin{tabular}[c]{@{}c@{}}0.64\\ (0.11)\end{tabular} \\
 & HuMAL-AL & \begin{tabular}[c]{@{}c@{}}0.56\\ (0.13)\end{tabular} & \begin{tabular}[c]{@{}c@{}}0.73\\ (0.06)\end{tabular} & \begin{tabular}[c]{@{}c@{}}0.54\\ (0.14)\end{tabular} & \begin{tabular}[c]{@{}c@{}}0.71*\\ (0.08)\end{tabular} \\
\bottomrule
\end{tabular}

\end{subtable}

\tiny
The results are the mean of 20 random bootstraps. ’*’ denotes a significant improvement over BERT model
\\ Yelp-50 denoted for reviews up to 50 words; Yelp-100 to 51-100 words, and Yelp-200 for reviews with length of 101-200 words.

\end{table*}

\subsection{Impact on BERT’s Attention}
We conducted a close analysis of attention allocation by BERT, \textit{HuMAL-AL}, and humans, aiming to address two pivotal questions: First, how do humans and BERT differ in their attention placement, particularly regarding the beginning versus the end of each sentence? Second, how does HuMAL change BERT’s attention allocation, and how does this change relate to human attention allocation? 

To ensure a ``fair" comparison, we converted the LLM's attention to a binary representation by assigning an attention value of 1 to words with significantly higher attention than the mean attention. According to our findings, humans tend to assign ``high attention" to approximately 13\% (myPersonality dataset) to 21\% (at Yelp task) of the words, while BERT's attention to such words is only about 5\%. Additionally, we conducted an alternative analysis using the interquartile range, which yielded the same insights. 

\begin{figure}[h]
  \begin{subfigure}{\columnwidth}
    \includegraphics[width=\textwidth]{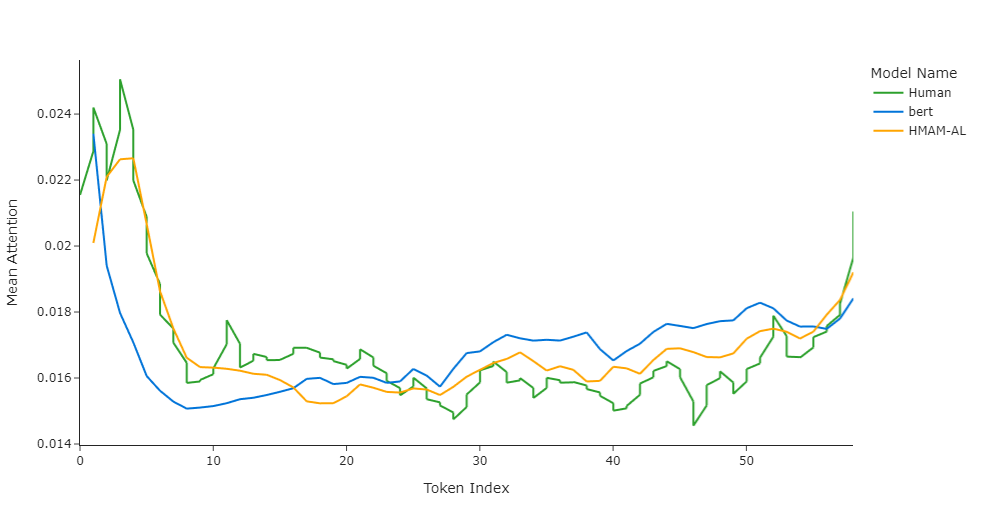}
    \\
    \subcaption{Yelp task}
    \label{fig:subfig1}
  \end{subfigure}
  \hfill
  \begin{subfigure}{\columnwidth}
    \includegraphics[width=\textwidth]{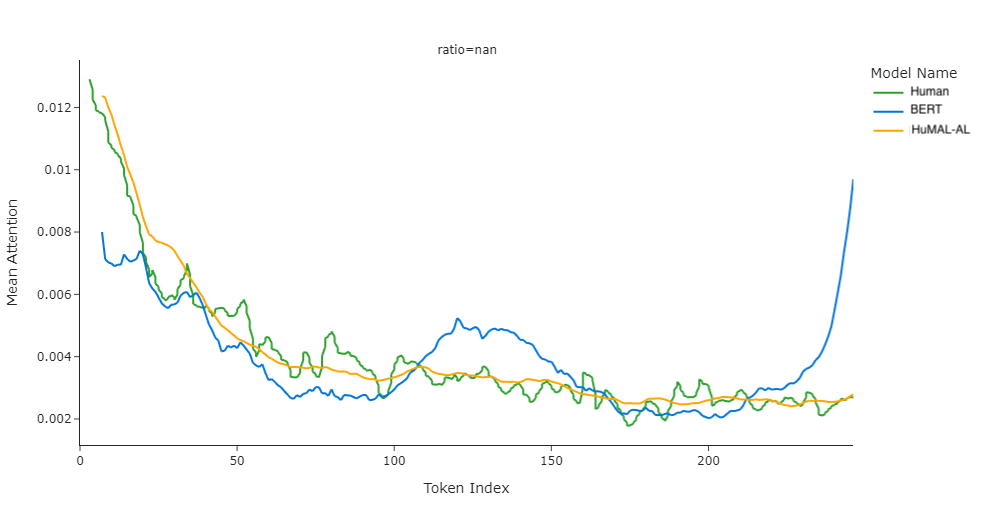}
    \subcaption{MyPersonality task}
    \label{fig:subfig2}
  \end{subfigure}
  \caption{Average attention as a function of token position}
  \label{fig:Binary_attention_per_token}
\end{figure}

Analyzing the mean attention assigned to a relative token position, as shown in Figure \ref{fig:Binary_attention_per_token}, shows that BERT’s attention starts high and then declines, with a W shape. Humans allocate attention more equally throughout the text, though they assign priority to the beginning of the text. HuMAL `smooths' the two curves, putting average attention in most of the tokens. It starts high and then decreases. 

Table \ref{tab:Mean_Cosine_Similarity}, shows a similarity analysis comparing the cosine similarity of human attention values to those of BERT and of HuMAL. As expected, HuMAL's implementation shifted the machine's attention closer to that of human annotators. Furthermore, HuMAL similarity scores are consistent in balanced datasets compared to unbalanced ones, while BERT's attention drifts away from that of humans as the dataset becomes more balanced.

\begin{table}[h]
    \centering
    \caption{Mean Cosine Similarity between Human Attention and Machine Attention (BERT and HuMAL)}
    \label{tab:Mean_Cosine_Similarity}
\centering
\centering
\begin{tabular}{lcccc}
\toprule
& \multicolumn{2}{c}{Label by Majority} & \multicolumn{2}{c}{Label by Self Report} \\ 
 & Imbalance & Balance & Imbalance & Balance \\
\midrule
BERT & \begin{tabular}[c]{@{}c@{}}0.27\\ (0.10)\end{tabular} & \begin{tabular}[c]{@{}c@{}}0.19\\ (0.02)\end{tabular} & \begin{tabular}[c]{@{}c@{}}0.23 \\ (0.11)\end{tabular} & \begin{tabular}[c]{@{}c@{}}0.19\\ (0.09)\end{tabular} \\
HuMAL-AL & \begin{tabular}[c]{@{}c@{}}0.45*\\ (0.02)\end{tabular} & \begin{tabular}[c]{@{}c@{}}0.47*\\ (0.02)\end{tabular} & \begin{tabular}[c]{@{}c@{}}0.45*\\ (0.03)\end{tabular} & \begin{tabular}[c]{@{}c@{}}0.48*\\ (0.02)\end{tabular} \\

\bottomrule
\end{tabular}

\end{table}

\subsection{Alternative Base Model}
We implemented \textit{HuMAL-AL} with two alternative LLMs---GPT-2 and XLNet---comparing their performance and the effectiveness of the HuMAL approach.

GPT-2, or Generative Pre-trained Transformer 2, is a prominent language model developed by OpenAI that operates on the transformer architecture. It is a generative model designed to predict the next word or sequence of words in a given context,In comparison to its successor, GPT-4, GPT-2 represents an earlier version with fewer parameters, indicating a subsequent evolution in model size and complexity. Furthermore, GPT-2 (unlike GPT-4) is open-source, allowing researchers and developers to access and build upon the model, fostering collaborative advancements in natural language processing \citep{radford2018improving}.

GPT-2, as its name suggests, is a generative text model. We added a classification head on top (linear layer) to turn it into a classification model. Based on the default provided by the Transformers library \citep{wolf2019huggingface}, it uses the last token to do the classification, as other causal models do. By design, the first token consistently receives the highest attention weight during model inference, yet paradoxically, this attention is found to be less informative \citep{vig2019analyzing}. Therefore we artificially add \(<|BeginningOfSentence|>\) tag, as advised in \citep{radford2018improving}.

XLNet integrates features from both encoder and decoder architectures. It adopts a permutation-based training objective, enabling it to capture a bidirectional context akin to BERT. Notably, during autoregressive text generation, XLNet employs a causal (left-to-right) self-attention mechanism, reminiscent of GPT models. This dual functionality positions XLNet as a versatile hybrid, serving as an encoder for pretraining purposes while functioning as a decoder for tasks involving text generation. The model's unique combination of bidirectional context capture and autoregressive capabilities contributes to its efficacy in a wide range of natural language processing applications.

Table \ref{alternative_performances} shows that GPT-2 and XLNET require fewer instances to train than BERT for the sentiment classification task. Yet, HuMAL significantly and economically improves performance with small datasets.

\begin{table*}[!ht]
\centering
\caption{HuMAL performance with alternative base models}
\label{alternative_performances}

\begin{subtable}{\textwidth}
\subcaption{Sentiment Classification Task}
\centering
\begin{tabular}{lcccccccccc}
\toprule
 &
  \multicolumn{5}{c}{Dataset Size = 250} &
  \multicolumn{5}{c}{Dataset Size = 500} \\
 &
  1\% &
  5\% &
  10\% &
  20\% &
  50\% &
  1\% &
  5\% &
  10\% &
  20\% &
  50\% \\ \midrule
GPT2 &
  \begin{tabular}[c]{@{}c@{}}0.58\\ (0.07)\end{tabular} &
  \begin{tabular}[c]{@{}c@{}}0.73\\ (0.07)\end{tabular} &
  \begin{tabular}[c]{@{}c@{}}0.81\\ (0.05)\end{tabular} &
  \begin{tabular}[c]{@{}c@{}}0.89\\ (0.04)\end{tabular} &
  \begin{tabular}[c]{@{}c@{}}0.95\\ (0.03)\end{tabular} &
  \begin{tabular}[c]{@{}c@{}}0.65\\ (0.08)\end{tabular} &
  \begin{tabular}[c]{@{}c@{}}0.88\\ (0.06)\end{tabular} &
  \begin{tabular}[c]{@{}c@{}}0.94\\ (0.02)\end{tabular} &
  \begin{tabular}[c]{@{}c@{}}0.97\\ (0.01)\end{tabular} &
  \begin{tabular}[c]{@{}c@{}}0.97\\ (0.01)\end{tabular} \\ 
HuMAL-AL-gpt &
  \begin{tabular}[c]{@{}c@{}}0.62\\ (0.07)\end{tabular} &
  \begin{tabular}[c]{@{}c@{}}0.79*\\ (0.06)\end{tabular} &
  \begin{tabular}[c]{@{}c@{}}0.87*\\ (0.07)\end{tabular} &
  \begin{tabular}[c]{@{}c@{}}0.89\\ (0.16)\end{tabular} &
  \begin{tabular}[c]{@{}c@{}}0.95\\ (0.02)\end{tabular} &
  \begin{tabular}[c]{@{}c@{}}0.73*\\ (0.06)\end{tabular} &
  \begin{tabular}[c]{@{}c@{}}0.89\\ (0.05)\end{tabular} &
  \begin{tabular}[c]{@{}c@{}}0.93\\ (0.06)\end{tabular} &
  \begin{tabular}[c]{@{}c@{}}0.97\\ (0.01)\end{tabular} &
  \begin{tabular}[c]{@{}c@{}}0.98\\ (0.01)\end{tabular} \\
xlnet &
  \begin{tabular}[c]{@{}c@{}}0.75\\ (0.1)\end{tabular} &
  \begin{tabular}[c]{@{}c@{}}0.93\\ (0.05)\end{tabular} &
  \begin{tabular}[c]{@{}c@{}}0.97\\ (0.02)\end{tabular} &
  \begin{tabular}[c]{@{}c@{}}0.98\\ (0.01)\end{tabular} &
  \begin{tabular}[c]{@{}c@{}}0.98\\ (0.01)\end{tabular} &
  \begin{tabular}[c]{@{}c@{}}0.8\\ (0.11)\end{tabular} &
  \begin{tabular}[c]{@{}c@{}}0.95\\ (0.05)\end{tabular} &
  \begin{tabular}[c]{@{}c@{}}0.97\\ (0.01)\end{tabular} &
  \begin{tabular}[c]{@{}c@{}}0.98\\ (0.02)\end{tabular} &
  \begin{tabular}[c]{@{}c@{}}0.98\\ (0.03)\end{tabular} \\ 
HuMAL3-xlnet &
  \begin{tabular}[c]{@{}c@{}}0.82*\\ (0.09)\end{tabular} &
  \begin{tabular}[c]{@{}c@{}}0.96*\\ (0.02)\end{tabular} &
  \begin{tabular}[c]{@{}c@{}}0.98\\ (0.01)\end{tabular} &
  \begin{tabular}[c]{@{}c@{}}0.98\\ (0.01)\end{tabular} &
  \begin{tabular}[c]{@{}c@{}}0.99\\ (0.01)\end{tabular} &
  \begin{tabular}[c]{@{}c@{}}0.9*\\ (0.03)\end{tabular} &
  \begin{tabular}[c]{@{}c@{}}0.95\\ (0.1)\end{tabular} &
  \begin{tabular}[c]{@{}c@{}}0.98\\ (0.02)\end{tabular} &
  \begin{tabular}[c]{@{}c@{}}0.99\\ (0.01)\end{tabular} &
  \begin{tabular}[c]{@{}c@{}}0.99*\\ (0.01)\end{tabular} \\

  \midrule
 &
  \multicolumn{5}{c}{Dataset Size = 1000} &
  \multicolumn{5}{c}{Dataset Size = 2000} \\
 &
  1\% &
  5\% &
  10\% &
  20\% &
  50\% &
  1\% &
  5\% &
  10\% &
  20\% &
  50\% \\ \midrule
GPT2 &
  \begin{tabular}[c]{@{}c@{}}0.74\\ (0.09)\end{tabular} &
  \begin{tabular}[c]{@{}c@{}}0.95\\ (0.02)\end{tabular} &
  \begin{tabular}[c]{@{}c@{}}0.97\\ (0.01)\end{tabular} &
  \begin{tabular}[c]{@{}c@{}}0.98\\ (0.01)\end{tabular} &
  \begin{tabular}[c]{@{}c@{}}0.98\\ (0.01)\end{tabular} &
  \begin{tabular}[c]{@{}c@{}}0.89\\ (0.05)\end{tabular} &
  \begin{tabular}[c]{@{}c@{}}0.98\\ (0.02)\end{tabular} &
  \begin{tabular}[c]{@{}c@{}}0.94\\ (0.14)\end{tabular} &
  \begin{tabular}[c]{@{}c@{}}0.98\\ (0.01)\end{tabular} &
  \begin{tabular}[c]{@{}c@{}}0.99\\ (0.01)\end{tabular} \\
HuMAL-AL-gpt &
  \begin{tabular}[c]{@{}c@{}}0.8*\\ (0.08)\end{tabular} &
  \begin{tabular}[c]{@{}c@{}}0.95\\ (0.02)\end{tabular} &
  \begin{tabular}[c]{@{}c@{}}0.97\\ (0.02)\end{tabular} &
  \begin{tabular}[c]{@{}c@{}}0.98\\ (0.01)\end{tabular} &
  \begin{tabular}[c]{@{}c@{}}0.98\\ (0.01)\end{tabular} &
  \begin{tabular}[c]{@{}c@{}}0.85*\\ (0.06)\end{tabular} &
  \begin{tabular}[c]{@{}c@{}}0.97\\ (0.02)\end{tabular} &
  \begin{tabular}[c]{@{}c@{}}0.98\\ (0.01)\end{tabular} &
  \begin{tabular}[c]{@{}c@{}}0.98\\ (0.01)\end{tabular} &
  \begin{tabular}[c]{@{}c@{}}0.99\\ (0.01)\end{tabular} \\
xlnet &
  \begin{tabular}[c]{@{}c@{}}0.84\\ (0.11)\end{tabular} &
  \begin{tabular}[c]{@{}c@{}}0.97\\ (0.03)\end{tabular} &
  \begin{tabular}[c]{@{}c@{}}0.98\\ (0.01)\end{tabular} &
  \begin{tabular}[c]{@{}c@{}}0.98\\ (0.02)\end{tabular} &
  \begin{tabular}[c]{@{}c@{}}0.99\\ (0.01)\end{tabular} &
  \begin{tabular}[c]{@{}c@{}}0.91\\ (0.09)\end{tabular} &
  \begin{tabular}[c]{@{}c@{}}0.97\\ (0.04)\end{tabular} &
  \begin{tabular}[c]{@{}c@{}}0.98\\ (0.01)\end{tabular} &
  \begin{tabular}[c]{@{}c@{}}0.98\\ (0.02)\end{tabular} &
  \begin{tabular}[c]{@{}c@{}}0.99\\ (0.01)\end{tabular} \\
HuMAL-AL-xlnet &
  \begin{tabular}[c]{@{}c@{}}0.95*\\ (0.03)\end{tabular} &
  \begin{tabular}[c]{@{}c@{}}0.98\\ (0.01)\end{tabular} &
  \begin{tabular}[c]{@{}c@{}}0.98\\ (0.01)\end{tabular} &
  \begin{tabular}[c]{@{}c@{}}0.99*\\ (0)\end{tabular} &
  \begin{tabular}[c]{@{}c@{}}0.99\\ (0.01)\end{tabular} &
  \begin{tabular}[c]{@{}c@{}}0.96*\\ (0.02)\end{tabular} &
  \begin{tabular}[c]{@{}c@{}}0.98\\ (0.01)\end{tabular} &
  \begin{tabular}[c]{@{}c@{}}0.99\\ (0.01)\end{tabular} &
  \begin{tabular}[c]{@{}c@{}}0.99*\\ (0.01)\end{tabular} &
  \begin{tabular}[c]{@{}c@{}}0.99*\\ (0.01)\end{tabular} \\

\bottomrule
\end{tabular}
\end{subtable}


\vspace{.5cm}

\begin{subtable}{\textwidth}
\centering
\subcaption{Personality Type Classification}
\centering
\begin{tabular}{lcccc}
\toprule
& \multicolumn{2}{c}{Label by Majority} & \multicolumn{2}{c}{Label by Self Report} \\ 
 & Imbalance & Balance & Imbalance & Balance \\
\midrule
GPT2 & \begin{tabular}[c]{@{}c@{}}0.53\\ (0.11)\end{tabular} & \begin{tabular}[c]{@{}c@{}}0.65\\ (0.10)\end{tabular} & \begin{tabular}[c]{@{}c@{}}0.49 \\ (0.06)\end{tabular} & \begin{tabular}[c]{@{}c@{}}0.60\\ (0.12)\end{tabular} \\
HuMAL-AL-gpt & \begin{tabular}[c]{@{}c@{}}0.52\\ (0.12)\end{tabular} & \begin{tabular}[c]{@{}c@{}}0.64\\ (0.13)\end{tabular} & \begin{tabular}[c]{@{}c@{}}0.52\\ (0.13)\end{tabular} & \begin{tabular}[c]{@{}c@{}}0.66*\\ (0.09)\end{tabular} \\
XLNET & \begin{tabular}[c]{@{}c@{}}0.53\\ (0.06)\end{tabular} & \begin{tabular}[c]{@{}c@{}}0.68\\ (0.07)\end{tabular} & \begin{tabular}[c]{@{}c@{}}0.58 \\ (0.09)\end{tabular} & \begin{tabular}[c]{@{}c@{}}0.74\\ (0.08)\end{tabular} \\
HuMAL-AL-xlnet & \begin{tabular}[c]{@{}c@{}}0.57*\\ (0.09)\end{tabular} & \begin{tabular}[c]{@{}c@{}}0.73*\\ (0.07)\end{tabular} & \begin{tabular}[c]{@{}c@{}}0.62\\ (0.08)\end{tabular} & \begin{tabular}[c]{@{}c@{}}0.80*\\ (0.05)\end{tabular} \\

\bottomrule
\end{tabular}

\end{subtable}

\tiny
The results are the mean of 20 random bootstraps. ’*’ denotes a significant improvement over the base model
\end{table*}

\section{Discussion and Conclusions}

This paper introduced and evaluated HuMAL, a novel approach for enhancing the performance of language models by aligning machine attention mechanisms with human attention mechanisms. Broadly, this approach incorporates human attention when training the machine by considering specific words or phrases that humans have annotated as relevant for their performance of a particular task. We show how this approach is implemented and when it is effective in boosting machine performance.. 

We explored three alternative strategies to incorporate human attention into an LLM’s attention mechanism: Attention as Loss (\textit{HuMAL-AL}), Attention as Normalizer (\textit{HuMAL-AN}), and Attention as Prior (\textit{HuMAL-AP}). We evaluated these strategies on two types of tasks: sentiment analysis, using a well-balanced annotated dataset of Yelp restaurant reviews \citep{sen2020human}; and personality-type classification (introvert vs. extrovert) based on social media posts (the myPersonality dataset \citep{stillwell2015mypersonality}), which we annotated for human attention. We compared our proposed HuMAL strategies to a baseline BERT model. Through experimentation, we found that regularization on the loss of the last attention layer (\textit{HuMAL-AL} strategy) yielded the best results, effectively aligning this layer to the human attention without increasing the classification loss. Moreover, \textit{HuMAL-AL} consistently outperformed BERT, as well as alternative LLMs (GPT-2 and XLNET), particularly in tasks with sparse and/or imbalanced training data.

Notably, our results seem to diverge from those of previous studies in which incorporating human attention into machine learning---through the use of sentiment lexicons \citep{zou2018lexicon} or cognitive data, such as eye-tracking and EEG \citep{mcguire2021sentiment}---did not lead to significant improvements in classification performance. One possible explanation for this disparity lies in the distinction between objective-aware and objective-unaware attention \citep{mccormick1997orienting}. Human attention can be characterized as either aware or unaware, referring to the extent to which individuals consciously focus their attention on specific aspects of the text. In normal, unaware reading, readers tend to allocate attention to terms that are more complex, longer, or ambiguous, reflecting their cognitive processing of the text. Conversely, in task-specific attention (e.g., sentiment classification), readers are more likely to search for cues or information directly related to the objective or task at hand. Our work concentrated on task-specific attention. It may be fruitful to investigate aligning machine attention to human unaware attention too.

We emphasize that the benefit of HuMAL over a base model is dependent on the task at hand and the availability of data. For example, LLMs such as GPT or Gemini are known to excel in relatively simple or well-documented tasks with abundant pre-trained data, such as text summarization, as well as sentiment analysis \citep{naveed2023comprehensive} -- one of the test cases we address. In contrast, humans generally outperform machines in zero-shot learning and in tasks that require intuitive understanding of context. In other words, machines may process information more quickly, but humans require fewer instances to learn effectively. Indeed, as exemplified herein, machines face challenges when the training set is extremely imbalanced or small in size, or when the task entails extracting nuanced aspects such as personality from text. Our findings suggest that HuMAL’s true potential lies in addressing such contexts. Examples of highly specialized tasks that currently require human expertise---and that might benefit from HuMAL---include detecting online drug dealers, identifying expert hackers, or understanding complex medical texts. Human experts could play a crucial role in guiding the machine learning process, ensuring that models can learn from the nuanced and often sparse data associated with these novel challenges \citep{te2023reciprocal}. 

Indeed, in the context of personality type classification, our approach illustrated how human attention can aid in capturing nuanced behavioral patterns associated with different personality traits. By incorporating these human insights into the model, we observed improved classification performance, reflecting an enhanced capacity to capture the subtleties of language that differentiate personality types.

It is important to acknowledge potential limitations in incorporating human attention into machine learning. First, though humans tend to exhibit more similarity to each other in their attention patterns than to machines, prior studies have identified substantial individual variability in human attention mechanisms. For example, \citet{mcguire2021sentiment}, who tracked attention using eye tracking and EEG data, had to average the attention of 12 readers due to significant variations in their attention patterns. Reassuringly, however, our annotation method, in which readers reported their own task-specific attention, produced a remarkable level of homogeneity, enabling us to suffice with just three annotators per instance---reducing the annotation burden and allowing us to achieve success where others have faced challenges. This outcome further highlights the benefits of HuMAL as a cost-effective alternative to more complex collection methods involving eye-tracking and brain imaging data, which require specialized equipment.

We further note that humans often lose focus when reading long texts, whereas machines benefit from having more signals—i.e., text—to learn from. This divergence in attention span and processing capacity underscores the complementary strengths of human and machine attention. By integrating human attention into machine models, as proposed with HuMAL, we can harness the depth of human insight while leveraging the computational power of machines to handle extensive textual data, ultimately achieving better performance in NLP tasks.

Another benefit of integrating human attention is the potential for enhanced interpretability of model decisions. Understanding the rationale behind a model's predictions is essential in many applications, particularly those involving high-stakes decisions. By aligning machine attention with human cognitive processes, the resulting models may be more transparent, enabling users to trace back and understand why certain decisions were made. This alignment also facilitates better communication between the model and human users, fostering trust and adoption of AI systems, without, however, compromising on machine performance that may be achieved with more advanced models that are low on interpretability \citep{te2023reciprocal}.

For future work, we plan to assess the performance of HuMAL on additional datasets and explore the use of other transformer-based models beyond BERT, such as LLAMA \citep{touvron2023llama}. Furthermore, we aim to conduct comprehensive linguistic analyses to better understand the distinctions in attention mechanisms between LLMs, HuMAL, and humans, specifically examining how each mechanism assigns attention to words based on their respective roles within sentences.

\bibliographystyle{plainnat}

\bibliography{references}

\begin{thebibliography}{36}
\providecommand{\natexlab}[1]{#1}
\providecommand{\url}[1]{\texttt{#1}}
\expandafter\ifx\csname urlstyle\endcsname\relax
  \providecommand{\doi}[1]{doi: #1}\else
  \providecommand{\doi}{doi: \begingroup \urlstyle{rm}\Url}\fi

\bibitem[Bao et~al.(2018)Bao, Chang, Yu, and Barzilay]{bao2018deriving}
Yujia Bao, Shiyu Chang, Mo~Yu, and Regina Barzilay.
\newblock Deriving machine attention from human rationales.
\newblock \emph{arXiv preprint arXiv:1808.09367}, 2018.

\bibitem[Bhardwaj et~al.(2023)Bhardwaj, Yang, and
  Cudre-Mauroux]{bhardwaj2023human}
Akansha Bhardwaj, Jie Yang, and Philippe Cudre-Mauroux.
\newblock Human-in-the-loop rule discovery for micropost event detection.
\newblock \emph{IEEE Transactions on Knowledge and Data Engineering},
  35\penalty0 (8):\penalty0 8100--8111, 2023.

\bibitem[Brauwers and Frasincar(2023)]{brauwers2021general}
Gianni Brauwers and Flavius Frasincar.
\newblock A general survey on attention mechanisms in deep learning.
\newblock \emph{IEEE Transactions on Knowledge and Data Engineering},
  35\penalty0 (4):\penalty0 3279--3298, 2023.

\bibitem[Clark et~al.(2019)Clark, Khandelwal, Levy, and Manning]{clark2019does}
Kevin Clark, Urvashi Khandelwal, Omer Levy, and Christopher~D Manning.
\newblock What does bert look at? an analysis of bert's attention.
\newblock \emph{arXiv preprint arXiv:1906.04341}, 2019.

\bibitem[Collins et~al.(2018)Collins, Rozanov, and
  Zhang]{collins2018evolutionary}
Edward Collins, Nikolai Rozanov, and Bingbing Zhang.
\newblock Evolutionary data measures: Understanding the difficulty of text
  classification tasks.
\newblock \emph{arXiv preprint arXiv:1811.01910}, 2018.

\bibitem[Cui et~al.(2021)Cui, Zhang, Che, Liu, and Chen]{cui2021understanding}
Yiming Cui, Wei-Nan Zhang, Wanxiang Che, Ting Liu, and Zhigang Chen.
\newblock Understanding attention in machine reading comprehension.
\newblock \emph{arXiv e-prints}, pages arXiv--2108, 2021.

\bibitem[Devlin et~al.(2018)Devlin, Chang, Lee, and Toutanova]{devlin2018bert}
Jacob Devlin, Ming-Wei Chang, Kenton Lee, and Kristina Toutanova.
\newblock Bert: Pre-training of deep bidirectional transformers for language
  understanding.
\newblock \emph{arXiv preprint arXiv:1810.04805}, 2018.

\bibitem[Guo et~al.(2021)Guo, Zhang, Liu, Zhu, Ballard, Hayhoe, and
  Stone]{guo2021machine}
Suna~Sihang Guo, Ruohan Zhang, Bo~Liu, Yifeng Zhu, Dana Ballard, Mary Hayhoe,
  and Peter Stone.
\newblock Machine versus human attention in deep reinforcement learning tasks.
\newblock \emph{Advances in Neural Information Processing Systems},
  34:\penalty0 25370--25385, 2021.

\bibitem[He et~al.(2019)He, Chen, Huang, Zhang, and Song]{he2019using}
Changai He, Sibao Chen, Shilei Huang, Jian Zhang, and Xiao Song.
\newblock Using convolutional neural network with bert for intent
  determination.
\newblock In \emph{2019 International Conference on Asian Language Processing
  (IALP)}, pages 65--70. IEEE, 2019.

\bibitem[Hollenstein et~al.(2018)Hollenstein, Rotsztejn, Troendle, Pedroni,
  Zhang, and Langer]{hollenstein2018zuco}
Nora Hollenstein, Jonathan Rotsztejn, Marius Troendle, Andreas Pedroni,
  Ce~Zhang, and Nicolas Langer.
\newblock Zuco, a simultaneous eeg and eye-tracking resource for natural
  sentence reading.
\newblock \emph{Scientific data}, 5\penalty0 (1):\penalty0 1--13, 2018.

\bibitem[Kim et~al.(2023)Kim, Mai, Han, Park, Nguyen, So, Singh, and
  Cha]{kim2023active}
Sundong Kim, Tung-Duong Mai, Sungwon Han, Sungwon Park, DK~Thi Nguyen, Jaechan
  So, Karandeep Singh, and Meeyoung Cha.
\newblock Active learning for human-in-the-loop customs inspection.
\newblock \emph{IEEE Transactions on Knowledge and Data Engineering},
  35\penalty0 (12):\penalty0 12039--12052, 2023.

\bibitem[Lei et~al.(2022)Lei, Zhang, Zhang, and Li]{lei2022coca}
Jiayu Lei, Zheng Zhang, Lan Zhang, and Xiang-Yang Li.
\newblock Coca: Cost-effective collaborative annotation system by combining
  experts and amateurs.
\newblock In \emph{2022 IEEE 38th International Conference on Data Engineering
  (ICDE)}, pages 674--685. IEEE, 2022.

\bibitem[Lei et~al.(2016)Lei, Barzilay, and Jaakkola]{lei2016rationalizing}
Tao Lei, Regina Barzilay, and Tommi Jaakkola.
\newblock Rationalizing neural predictions.
\newblock \emph{arXiv preprint arXiv:1606.04155}, 2016.

\bibitem[Lindsay(2020)]{lindsay2020attention}
Grace~W Lindsay.
\newblock Attention in psychology, neuroscience, and machine learning.
\newblock \emph{Frontiers in computational neuroscience}, 14:\penalty0 516985,
  2020.

\bibitem[Liu et~al.(2017)Liu, Mao, Sha, and Yuille]{liu2017attention}
Chenxi Liu, Junhua Mao, Fei Sha, and Alan Yuille.
\newblock Attention correctness in neural image captioning.
\newblock In \emph{Proceedings of the AAAI conference on artificial
  intelligence}, volume~31, 2017.

\bibitem[Luong et~al.(2015)Luong, Pham, and Manning]{luong2015effective}
Minh-Thang Luong, Hieu Pham, and Christopher~D Manning.
\newblock Effective approaches to attention-based neural machine translation.
\newblock \emph{arXiv preprint arXiv:1508.04025}, 2015.

\bibitem[Makino et~al.(2022)Makino, Jastrzkebski, Oleszkiewicz, Chacko,
  Ehrenpreis, Samreen, Chhor, Kim, Lee, Pysarenko,
  et~al.]{makino2022differences}
Taro Makino, Stanislaw Jastrzkebski, Witold Oleszkiewicz, Celin Chacko, Robin
  Ehrenpreis, Naziya Samreen, Chloe Chhor, Eric Kim, Jiyon Lee, Kristine
  Pysarenko, et~al.
\newblock Differences between human and machine perception in medical
  diagnosis.
\newblock \emph{Scientific reports}, 12\penalty0 (1):\penalty0 6877, 2022.

\bibitem[McCormick(1997)]{mccormick1997orienting}
Peter~A McCormick.
\newblock Orienting attention without awareness.
\newblock \emph{Journal of Experimental Psychology: Human Perception and
  Performance}, 23\penalty0 (1):\penalty0 168, 1997.

\bibitem[McGuire and Tomuro(2021)]{mcguire2021sentiment}
Erik McGuire and Noriko Tomuro.
\newblock Sentiment analysis with cognitive attention supervision.
\newblock In \emph{Canadian Conference on AI}, 2021.

\bibitem[Naveed et~al.(2023)Naveed, Khan, Qiu, Saqib, Anwar, Usman, Akhtar,
  Barnes, and Mian]{naveed2023comprehensive}
Humza Naveed, Asad~Ullah Khan, Shi Qiu, Muhammad Saqib, Saeed Anwar, Muhammad
  Usman, Naveed Akhtar, Nick Barnes, and Ajmal Mian.
\newblock A comprehensive overview of large language models.
\newblock \emph{arXiv preprint arXiv:2307.06435}, 2023.

\bibitem[Petricek et~al.(2023)Petricek, van Den~Burg, Naz{\'a}bal, Ceritli,
  Jim'enez-Ruiz, and Williams]{petricek2023ai}
Tomas Petricek, Gerrit~JJ van Den~Burg, Alfredo Naz{\'a}bal, Taha Ceritli,
  Ernesto Jim'enez-Ruiz, and Christopher~KI Williams.
\newblock Ai assistants: A framework for semi-automated data wrangling.
\newblock \emph{IEEE Transactions on Knowledge and Data Engineering},
  35\penalty0 (9):\penalty0 9295--9306, 2023.

\bibitem[Radford et~al.(2018)Radford, Narasimhan, Salimans, and
  Sutskever]{radford2018improving}
Alec Radford, Karthik Narasimhan, Tim Salimans, and Ilya Sutskever.
\newblock Improving language understanding by generative pre-training, 2018.

\bibitem[Radford et~al.(2019)Radford, Wu, Child, Luan, Amodei, and
  Sutskever]{radford2019language}
Alec Radford, Jeffrey Wu, Rewon Child, David Luan, Dario Amodei, and Ilya
  Sutskever.
\newblock Language models are unsupervised multitask learners.
\newblock \emph{OpenAI blog}, 1\penalty0 (8):\penalty0 9, 2019.

\bibitem[Sen et~al.(2020)Sen, Hartvigsen, Yin, Kong, and
  Rundensteiner]{sen2020human}
Cansu Sen, Thomas Hartvigsen, Biao Yin, Xiangnan Kong, and Elke Rundensteiner.
\newblock Human attention maps for text classification: Do humans and neural
  networks focus on the same words?
\newblock In \emph{Proceedings of the 58th annual meeting of the association
  for computational linguistics}, pages 4596--4608, 2020.

\bibitem[Stillwell and Kosinski(2015)]{stillwell2015mypersonality}
DJ~Stillwell and M~Kosinski.
\newblock mypersonality project website, 2015.

\bibitem[Te’eni et~al.(2023)Te’eni, Yahav, Zagalsky, Schwartz, Silverman,
  Cohen, Mann, and Lewinsky]{te2023reciprocal}
Dov Te’eni, Inbal Yahav, Alexely Zagalsky, David Schwartz, Gahl Silverman,
  Daniel Cohen, Yossi Mann, and Dafna Lewinsky.
\newblock Reciprocal human-machine learning: A theory and an instantiation for
  the case of message classification.
\newblock \emph{Management Science}, 2023.

\bibitem[Touvron et~al.(2023)Touvron, Lavril, Izacard, Martinet, Lachaux,
  Lacroix, Roziere, Goyal, Hambro, Azhar, et~al.]{touvron2023llama}
Hugo Touvron, Thibaut Lavril, Gautier Izacard, Xavier Martinet, Marie-Anne
  Lachaux, Timothee Lacroix, Baptiste Roziere, Naman Goyal, Eric Hambro, Faisal
  Azhar, et~al.
\newblock Llama: Open and efficient foundation language models.
\newblock \emph{arXiv preprint arXiv:2302.13971}, 2023.

\bibitem[Vania and Lopez(2017)]{vania2017characters}
Clara Vania and Adam Lopez.
\newblock From characters to words to in between: Do we capture morphology?
\newblock \emph{arXiv preprint arXiv:1704.08352}, 2017.

\bibitem[Vaswani et~al.(2017)Vaswani, Shazeer, Parmar, Uszkoreit, Jones, Gomez,
  Kaiser, and Polosukhin]{vaswani2017attention}
Ashish Vaswani, Noam Shazeer, Niki Parmar, Jakob Uszkoreit, Llion Jones,
  Aidan~N Gomez, Lukasz Kaiser, and Illia Polosukhin.
\newblock Attention is all you need.
\newblock In \emph{Advances in neural information processing systems}, pages
  5998--6008, 2017.

\bibitem[Vig and Belinkov(2019)]{vig2019analyzing}
Jesse Vig and Yonatan Belinkov.
\newblock Analyzing the structure of attention in a transformer language model.
\newblock \emph{arXiv preprint arXiv:1906.04284}, 2019.

\bibitem[Wolf et~al.(2019)Wolf, Debut, Sanh, Chaumond, Delangue, Moi, Cistac,
  Rault, Louf, Funtowicz, et~al.]{wolf2019huggingface}
Thomas Wolf, Lysandre Debut, Victor Sanh, Julien Chaumond, Clement Delangue,
  Anthony Moi, Pierric Cistac, Tim Rault, R'emi Louf, Morgan Funtowicz, et~al.
\newblock Huggingface's transformers: State-of-the-art natural language
  processing.
\newblock \emph{arXiv preprint arXiv:1910.03771}, 2019.

\bibitem[Yoo et~al.(2022)Yoo, Rosenberg, Kwon, Lin, Avery, Sheinost, Constable,
  and Chun]{yoo2022brain}
Kwangsun Yoo, Monica~D Rosenberg, Young~Hye Kwon, Qi~Lin, Emily~W Avery, Dustin
  Sheinost, R~Todd Constable, and Marvin~M Chun.
\newblock A brain-based general measure of attention.
\newblock \emph{Nature human behaviour}, 6\penalty0 (6):\penalty0 782--795,
  2022.

\bibitem[Zagalsky et~al.(2021)Zagalsky, Te'eni, Yahav, Schwartz, Silverman,
  Cohen, Mann, and Lewinsky]{zagalsky2021design}
Alexey Zagalsky, Dov Te'eni, Inbal Yahav, David~G Schwartz, Gahl Silverman,
  Daniel Cohen, Yossi Mann, and Dafna Lewinsky.
\newblock The design of reciprocal learning between human and artificial
  intelligence.
\newblock \emph{Proceedings of the ACM on Human-Computer Interaction},
  5\penalty0 (CSCW2):\penalty0 1--36, 2021.

\bibitem[Zhao et~al.(2023)Zhao, Zhang, and Zhang]{zhao2023does}
Lei Zhao, Yingyi Zhang, and Chengzhi Zhang.
\newblock Does attention mechanism possess the feature of human reading? a
  perspective of sentiment classification task.
\newblock \emph{Aslib Journal of Information Management}, 75\penalty0
  (1):\penalty0 20--43, 2023.

\bibitem[Zou et~al.(2023)Zou, Zhang, Li, Tian, and Ding]{zou2023human}
Jiajie Zou, Yuran Zhang, Jialu Li, Xing Tian, and Nai Ding.
\newblock Human attention during goal-directed reading comprehension relies on
  task optimization.
\newblock \emph{bioRxiv}, pages 2023--04, 2023.

\bibitem[Zou et~al.(2018)Zou, Gui, Zhang, and Huang]{zou2018lexicon}
Yicheng Zou, Tao Gui, Qi~Zhang, and Xuan-Jing Huang.
\newblock A lexicon-based supervised attention model for neural sentiment
  analysis.
\newblock In \emph{Proceedings of the 27th international conference on
  computational linguistics}, pages 868--877, 2018.

\end{thebibliography}

\end{document}